\documentclass[9pt,twocolumn,twoside]{osajnl}

\usepackage{times}

\usepackage{amsmath}
\usepackage{amssymb}
\usepackage{multirow}
\usepackage{multirow}
\usepackage{mathtools}
\usepackage{subfloat}
\usepackage{float}
\usepackage{xcolor}
\usepackage{url}
\usepackage{bbm}
\usepackage{algorithm}
\usepackage{algpseudocode}
\usepackage{pbox}
\usepackage{placeins}
\usepackage{bm}
\usepackage{acronym}
\usepackage{utfsym}
\usepackage{rotating}
\usepackage{siunitx}
\usepackage{pifont}

\journal{ol} 

\setboolean{shortarticle}{false}

\title{Scientific Exploration of Challenging Planetary Analog Environments with a Team of Legged Robots}

\author[1,+,*]{Philip Arm}
\author[1,+]{Gabriel Waibel}
\author[1]{Jan Preisig}
\author[1]{Turcan Tuna}
\author[1,2]{Ruyi Zhou}
\author[3,4]{Valentin Bickel}
\author[5]{Gabriela Ligeza}
\author[1]{Takahiro Miki}
\author[6,7,8]{Florian Kehl}
\author[1]{Hendrik Kolvenbach}
\author[1]{Marco Hutter}

\affil[1]{Robotic Systems Lab, ETH Zurich, Zurich, Switzerland}
\affil[2]{State Key Laboratory of Robotics and System, Harbin Institute of Technology, Harbin, China}
\affil[3]{Laboratory of Hydraulics, Hydrology and Glaciology, ETH Zurich, Zurich, Switzerland}
\affil[4]{Center for Space and Habitability, University of Bern, Bern, Switzerland}
\affil[5]{Department of Environmental Sciences, University of Basel, Basel, Switzerland}
\affil[6]{Innovation Cluster Space and Aviation (UZH Space Hub), University of Zurich, Dübendorf, Switzerland}
\affil[7]{Center for Theoretical Astrophysics and Cosmology, University of Zurich, Zurich, Switzerland}
\affil[8]{Institute of Medical Engineering, Space Biology Group, Lucerne University of Applied Sciences and Art, Hergiswil, Switzerland}
\affil[+]{These authors contributed equally to this work.}
\affil[*]{Corresponding author: parm@ethz.ch \newline

This is the accepted version of Science Robotics Vol. 8, Issue 80, eade9548 (2023)
DOI: 10.1126/scirobotics.ade9548}

\begin{abstract}
The interest in exploring planetary bodies for scientific investigation and in-situ resource utilization is ever-rising. Yet, many sites of interest are inaccessible to state-of-the-art planetary exploration robots because of the robots' inability to traverse steep slopes, unstructured terrain, and loose soil. Additionally, current single-robot approaches only allow a limited exploration speed and a single set of skills. Here, we present a team of legged robots with complementary skills for exploration missions in challenging planetary analog environments. We equipped the robots with an efficient locomotion controller, a mapping pipeline for online and post-mission visualization, instance segmentation to highlight scientific targets, and scientific instruments for remote and in-situ investigation. Furthermore, we integrated a robotic arm on one of the robots to enable high-precision measurements. Legged robots can swiftly navigate representative terrains, such as granular slopes beyond 25 degrees, loose soil, and unstructured terrain, highlighting their advantages compared to wheeled rover systems. We successfully verified the approach in analog deployments at the BeyondGravity ExoMars rover testbed, in a quarry in Switzerland, and at the Space Resources Challenge in Luxembourg. Our results show that a team of legged robots with advanced locomotion, perception, and measurement skills, as well as task-level autonomy, can conduct successful, effective missions in a short time. Our approach enables the scientific exploration of planetary target sites that are currently out of human and robotic reach.
\end{abstract}

\begin{document}

\maketitle

\section*{Introduction}
Robotic planetary exploration is invaluable for advancing our understanding of the solar system and enabling the prospection of potential resources. The recent commitment of national and commercial entities to return to the Moon - targeting a sustainable, long-term human presence - boosted the development of robotic exploration technologies. 

Many science-, exploration-, and resource extraction-relevant targets across the lunar surface lie in hard-to-reach areas or areas with substantial potential to host unknown physical surface properties. Examples include pyroclastic vents, volcanic rilles, caves, irregular mare patches, and fresh impact craters \cite{qiao2021ina,glotch2021scientific}. Consequently, developing robotic exploration systems that can efficiently traverse challenging terrain without compromising their explorative, scientific, and resource prospection capabilities remains a top priority.

Notably, several lunar exploration efforts revolve around the National Aeronautics and Space Administration's (NASA) Artemis program, which focuses on robotic and crewed science and exploration at the lunar south pole \cite{smith2020artemis}. One of the first Artemis program missions, the Volatiles Investigating Polar Exploration Rover (VIPER) \cite{colaprete2021volatiles}, will venture into several permanently shadowed regions - cold, volatile-rich topographic depressions that have not been illuminated for millions of years \cite{colaprete2010detection,mitrofanov2010hydrogen,li2018direct}. Ultimately, Artemis 3 will lead humans to the south pole in 2025 \cite{scoville2022artemis}. All of those missions will need to navigate the challenging south-polar terrain, including steep slopes, impact ejecta and boulder fields, and potentially anomalous physical regolith properties \cite{flahaut2020regions,spudis2008geology,colaprete2021volatiles}. 

One promising way to foster the development of lunar exploration and prospection technologies is challenge-driven innovation. The European Space Agency (ESA) and the European Space Resources Innovation Centre (ESRIC) established the Space Resources Challenge (SRC) in 2021 to evaluate and advance the state of the art of robotic lunar prospecting technologies. The main technical goal of the challenge was prospecting a lunar analog environment for resource-enriched areas (REAs), meaning areas that contain minerals suitable for In Situ Resource Utilization (ISRU), such as ilmenite, rutile, and titanium dioxide. The two rounds of the challenge took place in 2021 and 2022, with the final competition in a lunar analog terrain in Luxembourg. The competition involved adverse conditions found at the lunar south pole, including a previously unknown terrain, loose granular soil, high solar incidence angle illumination creating long, high-contrast shadows, and network communications with high latency 5.0 s round trip time (RTT)) and intermittent complete loss of signal. The SRC was an important inspiration in this work and was one of the two major field deployments of our robotic exploration team.

In this article, we propose a team of legged robots for quick, efficient, and safe exploration and prospection of challenging planetary analog environments. The team approach allows us to cover a larger area, deploy a wider variety of scientific payloads, investigate more scientific targets, and gain more in-depth knowledge per target than possible with a single-robot approach or non-teamed multi-robot approaches. Additionally, the increased redundancy allows mission completion even if multiple robots fail. To control the robotic team, the operators send high-level navigation, remote measurement, and in-situ measurement tasks to the robots. The robots execute these tasks autonomously using their state-of-the-art mobility and navigation systems, as well as a complementary and redundant set of payloads. The level of autonomy allows continued scientific data collection, even if communication becomes unreliable or during a complete loss of signal (LoS). Simultaneously, the scientists in the operations team can select and prioritize scientific targets during the mission.

Until now, most planetary exploration robots relied on wheeled locomotion. Their locomotion system did not fundamentally change since the first rover - Lunokhod 1 - touched down on the surface of the Moon in 1970 \cite{carrier1991physical}. Other prominent examples are the Lunar Roving Vehicle (LRV), Lunokhod 2, and Yutu 2 \cite{costes1972mobility,carrier1991physical,florenskii1978floor,ding20222}, or martian rovers such as the Sojourner Rover \cite{jones1997really}, Spirit, Opportunity \cite{lindemann2006mars},  Curiosity \cite{grotzinger2012mars}, and Perseverance \cite{farley2020mars}. Although these systems can build upon well-tested heritage technology and provide robustness in relatively flat terrain, wheeled rovers reach their limitations on steep slopes, on loose granular terrain, and in unstructured environments. On Mars, the Spirit rover was lost in anomalously loose soil \cite{webster2009nasa} and Opportunity got temporarily stuck in a dune \cite{david2005opportunity}. On the Moon, the Apollo 15 LRV was trapped in loose regolith and had to be manually retrieved by the astronauts \cite{costes1972mobility}. Similarly, Lunokhod 2 encountered excessive wheel sinkage ($>$20 cm) near Le Monnier crater \cite{florenskii1978floor}. The Yutu-2 team reported that entering craters would be of great scientific interest. However, they do not target craters because of the increased probability of locomotion failure \cite{ding20222}. This locomotion limitation prevents current missions from investigating high-priority targets \cite{potts2015,steenstra2016,seeni2010,qiao2021ina,glotch2021scientific}.

Meanwhile, terrestrial legged robots have reached a high level of robustness in exploring unknown environments. Their robust locomotion system allows them to traverse unstructured, challenging natural terrain, including mud, gravel, snow, vegetation, and sand \cite{lee2020learning,miki2022learning}. 

Several researchers have developed legged robots with the intent to use them in space in the past \cite{roennau2014,dirk2007bio,bartsch2012spaceclimber,roennau2014reactive}. Until now, we have focused on the usage of dynamically walking legged robots on steep, planetary soil analogs \cite{kolvenbach2021martianslopes} and low-gravity environments \cite{kolvenbach2018isairas,kolvenbach2019iros,rudin2021catlike}, showcasing the potential of the technology. However, for these robots to be useful in real-world scenarios, they need to be advanced beyond locomotion tasks. They have to interact with their environment in realistic analog missions, for example by deploying scientific instruments or taking samples. We advanced in this direction at the first field trial of the SRC, where we deployed a legged robot with base-mounted instruments \cite{arm2022results}.

Heterogeneous robotic teams have been used as a viable solution in terrestrial real-world missions. Notably, all top-ranking teams in the DARPA Subterranean Challenge 2021 used heterogeneous robotic teams with diverse skills \cite{tranzatto2022cerberus,hudson2021heterogeneous,agha2021nebula}. To succeed in the challenge, the teams developed robust solutions for locomotion, localization, multi-robot mapping, local planning, and exploration planning. In this work, we build upon these advances - specifically on the systems of our team CERBERUS~\cite{tranzatto2022cerberus} - and address the unique challenges presented by analog space missions, including instrument deployment, efficient and robust locomotion with robotic arms, redundancy to component or system failures, and validation in realistic missions with high latency communication. Additionally, although the robotic teams in the Subterranean Challenge were diverse in their locomotion skills, we consider diversity in scientific investigation capabilities in this work.

Several robotic teams for planetary exploration have been developed and tested in analog environments. For example, the German Aerospace Center (DLR) deployed a drone and two wheeled robots autonomously to set up a distributed radio telescope and perform geological exploration on Mt. Etna \cite{schuster2020arches}. Although they showed a high level of autonomy, the wheeled rovers were limited in their locomotion capabilities. The German Research Centre for Artificial Intelligence (DFKI) developed a heterogeneous robotic team of a wheeled and a legged robot for sample collection in a lunar analog environment \cite{cordes2011lunares}. In \cite{sonsalla2017field}, the authors demonstrate a teleoperated sample return mission using a robotic team in a martian analog environment. The operations team operated the robots via waypoints and tested capabilities such as sampling in isolated tests instead of the full mission deployment. More recently, NASA JPL built upon their NEBULA solution to explore analog martian caves with multiple Spot robots in the NASA BRAILLE project \cite{nasa2022braille}. One of the robots was equipped with a robotic arm to take close-up images and swab samples. However, the details of this work are not yet published. Lastly, the first heterogeneous robotic team for planetary exploration is currently operating on Mars: The Ingenuity helicopter supports the Mars 2020 mission by scouting potential targets for the Perseverance rover and inspecting targets the rover cannot access \cite{tzanetos2022ingenuity}. This actual Mars mission is a remarkable example of a heterogeneous robotic team for planetary exploration.

We present our teamed exploration approach with dynamically walking robots for planetary environments (Movie~1). We designed a team of legged robots with a diverse set of scientific investigation skills and redundancy measures and validated our system in three challenging analog environments: The ExoMars locomotion test facility, a quarry site, and the competition site of the SRC. We report our results and lessons learned from these deployments and identify opportunities for future developments.

To achieve effective analog mission deployments, we developed, improved and validated critical subsystems: We validated the baseline locomotion policy \cite{miki2022learning} on planetary analog terrain and developed a locomotion policy with a focus on efficiency. Our policy includes arm observations, making it robust for legged robots with robotic arms for scientific investigation in challenging environments. Additionally, we built a two-pronged mapping approach for lightweight real-time online mapping and high-resolution, realistic post-mission visualization. Our instance segmentation pipeline highlights potential scientific targets to support online mission planning in previously unknown environments. By distributing a balanced set of remote and close-up scientific instruments, we achieve effective and safe exploration missions.

We showcase the capabilities of our legged robots in martian and lunar analog environments, demonstrating that our technology can enable robots to investigate scientifically transformative targets on the Moon and Mars that are unreachable at present using wheeled rover systems. 

\begin{figure*}
    \centering
    \includegraphics[width=0.9\textwidth]{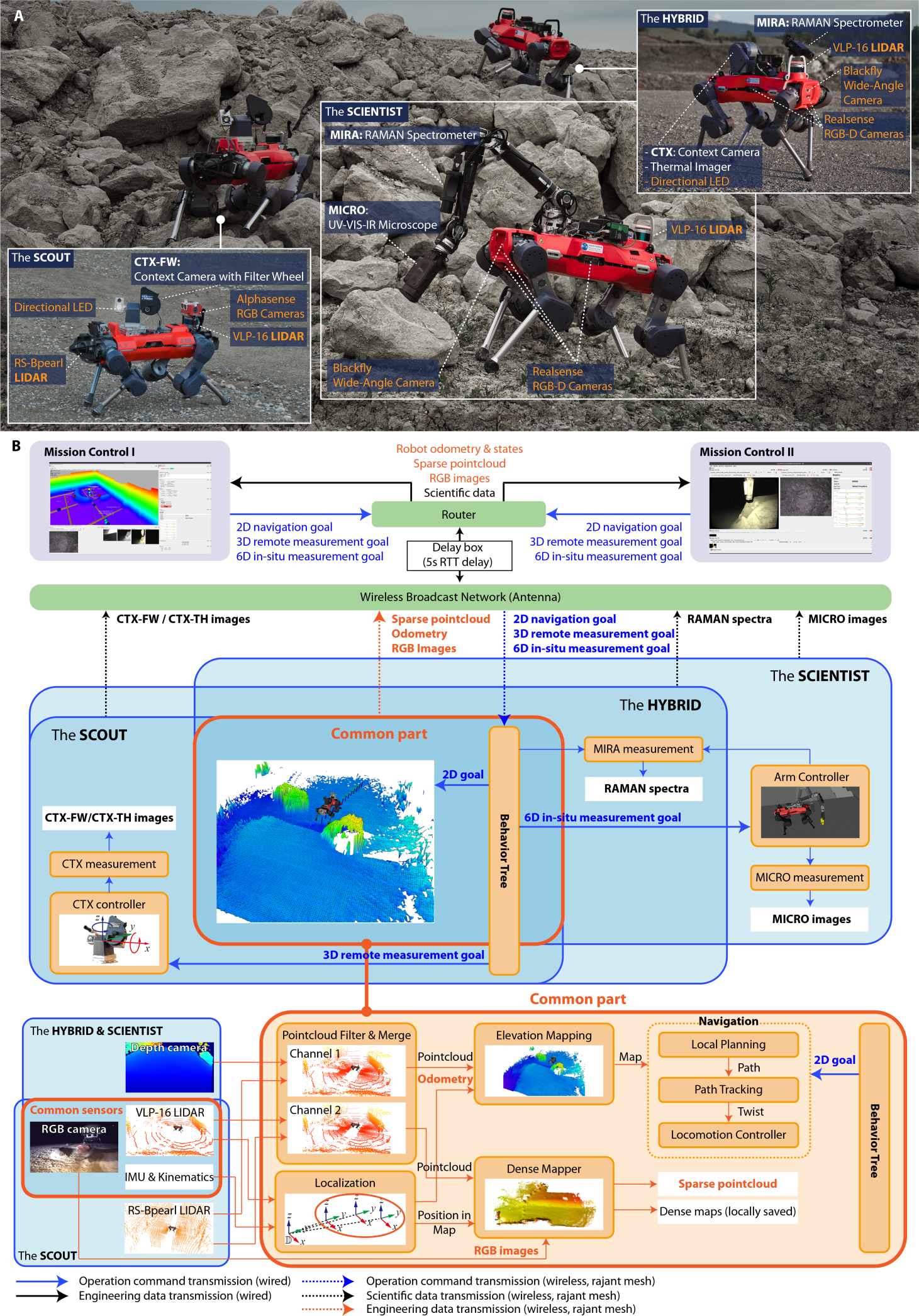}
    \caption{\textbf{System architecture of our team of legged robots.} (\textbf{A})  Robotic and scientific payloads on the Scout, Scientist, and Hybrid. Robotic payloads and scientific payloads are labeled in orange and white, respectively. (\textbf{B}) High-level overview of the software architecture of our system. With a balanced combination of shared and specialized modules per robot, we designed a safe yet efficient multirobot system.}
\end{figure*}

\begin{figure*}
    \centering
    \includegraphics[width=\textwidth]{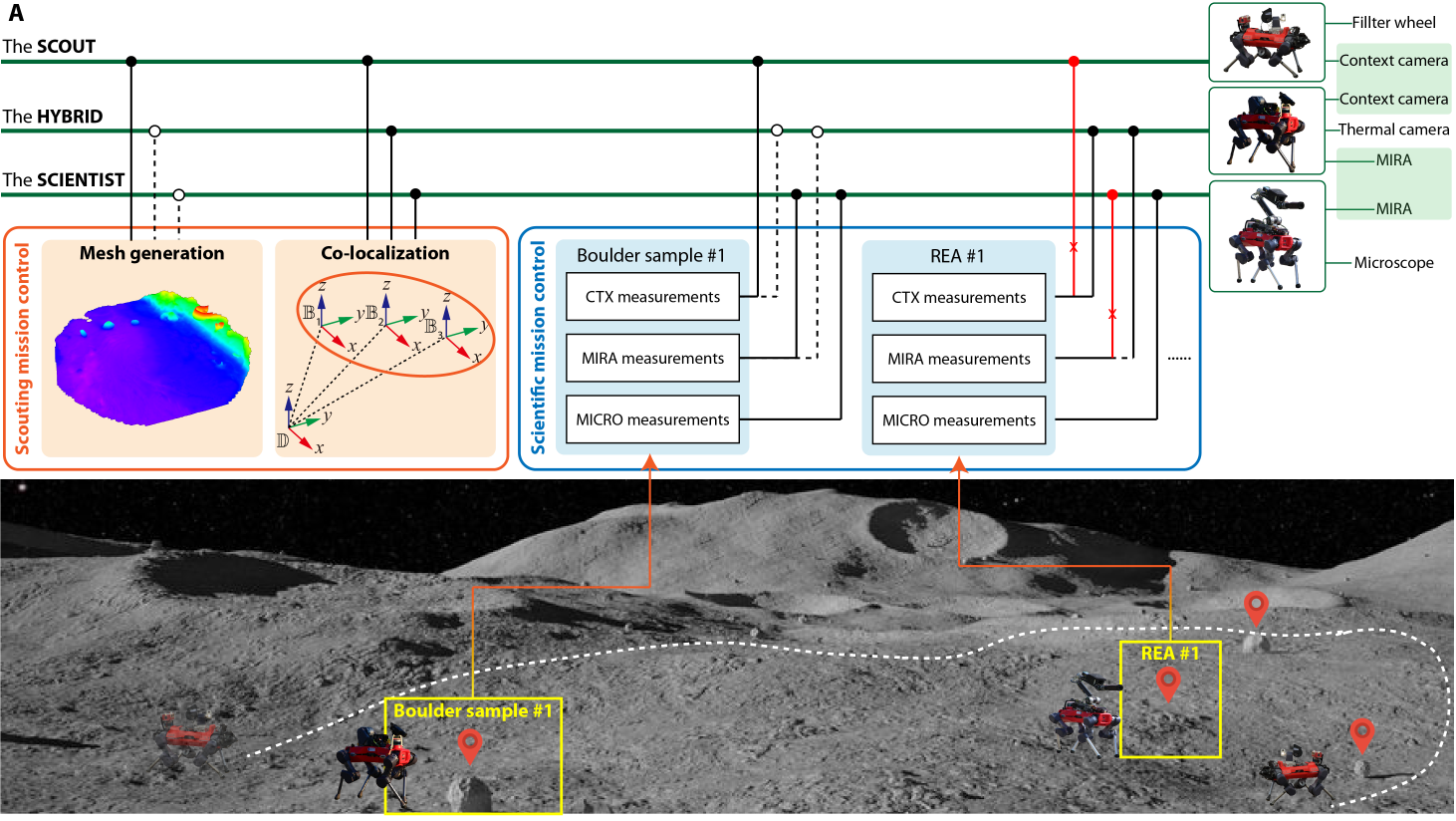}
    \caption{\textbf{Example of task allocations in our system.} Task allocations are shown as solid lines. Alternative allocation paths are shown as dashed lines. When a task allocation becomes invalid, for example, because a payload or robot malfunctions (red lines), tasks can be reallocated according to our redundancy concept.}
\end{figure*}

\section*{Results}

\subsection*{A Team of Legged Robots for Planetary Exploration}
The core of our system comprises a team of three four-legged robots. ANYmal \cite{hutter2017anymal} serves as the base platform for all robots. Each of the three robots has a dedicated role and a unique set of payloads for exploration and scientific data collection. Fig.~1-A presents our team of legged robots: the Scout, the Hybrid, and the Scientist. The Scout's primary task is rapidly exploring the environment using its additional LiDARs and RGB cameras. It provides the operations team, consisting of the robot operators and planetary scientists, with an overview of the previously unknown area and allows the team to identify potential scientific targets. Its secondary task is to capture images of potential scientific targets in various spectral bands using a pan-tilt context imager augmented with a custom-built filter wheel (CTX-FW) (Fig.~S1). The Hybrid's main task is collecting scientific data of numerous targets using a pan-tilt context imager with an additional thermal camera (CTX-TH). Moreover, it is equipped with a base-mounted Metrohm Instant Raman Analyzer XTR DS (MIRA) with a zoom lens to acquire Raman spectra of targets of interest. The Scientist performs an in-depth scientific analysis of previously identified targets. It features a custom 6-degrees-of-freedom (DoF) robotic arm with a MIRA on the forearm and a custom microscopic imager (MICRO) on the wrist.

Fig.~1-B shows a system overview. Two operators on two mission control stations send high-level navigation, remote measurement, and in-situ measurement goals to the robots. "Remote measurement tasks" describe scientific tasks that the robot can conduct remotely using the CTX-FW and the CTX-TH payloads. "In-situ measurement tasks" describe close-up investigations, namely MIRA and MICRO measurements. All data packets between mission control and the robots are delayed with a round-trip-time (RTT) of 5.0 s to simulate lunar operation. Navigation goals are handled by the same module on each robot. 3D remote measurement goals are only used by the CTX imagers on the Scout and the Hybrid. The 6D in-situ measurement targets are processed by the Hybrid and the Scientist.
The robots send feedback about their state, navigation images, a sparse map representation, and scientific data of targets of interest to the mission control stations.

Although all robots have their designated role, they share many exploration capabilities and payloads, enabling a high redundancy level. If a robot fails, the operations team can reallocate tasks between the robots. Fig.~2 visualizes how tasks can be reallocated for examples of exploration and measurement tasks thanks to the payload redundancy concept. Table~S1 summarizes which scientific measurement tasks can be executed under which failure conditions.

\subsection*{Experimental Setup on Analog Sites}
We conducted end-to-end missions in two lunar analog environments and a locomotion validation test in a martian analog testbed. In both end-to-end analog missions, the operations team consisted of five people: A team of one robot operator and one planetary scientist each operated one of the two mission control stations. The robot operators sent tasks to the robots, and the planetary scientists selected and prioritized targets based on the data received from the robots. Additionally, one supervisor overlooked the mission and ensured communication between the two control teams. Both mission control stations could be used to interact with any of the robots. However, the team ensured that each robot only received commands from one control team at any given time to prevent conflicting commands, where the newer command would supersede the previous one. In the presented deployments, we used one mission control station to control the Scout and the Hybrid and the second control station to control the Scientist.

\subsubsection*{Space Resources Challenge, Luxembourg}
One field campaign occurred at the ESA/ESRIC Space Resources Challenge in Esch-sur-Alzette, Luxemburg (September 2022). The competition area measured 1800 m$^2$, and the terrain was unknown before the challenge. The ground was covered in coarse, granular basalt with a substantial fine fraction (clay to silt). The light conditions closely resembled those at the lunar south pole because of a powerful illumination source at a high incidence angle in the corner of the area (Fig.~3-A). The analog scenario contained several locations of interest, such as resource-enriched areas, boulders, craters, and a lunar habitat prototype. The mission control room was separated from the competition area with a 5.0s RTT to communicate to all systems in the competition area. We summarize the most important objectives and rules of the SRC in Table~S2.

\begin{figure*}
    \centering
    \includegraphics[width=1.0\textwidth]{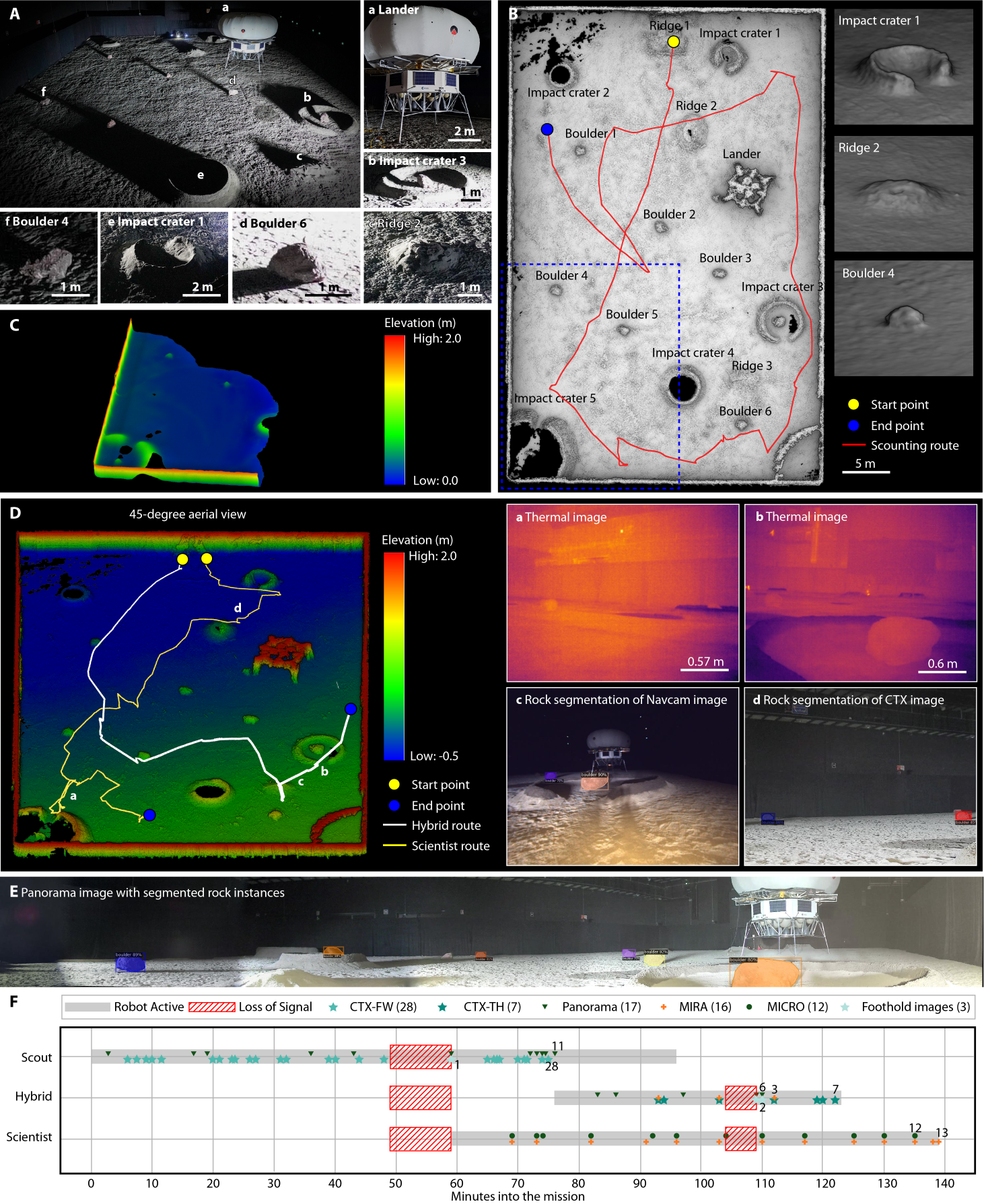}
    \caption{\textbf{Performance of our team of legged robots during the SRC.} (\textbf{A})  Overview of the competition area. (\textbf{B}) Postprocessed high-resolution map of the exploration area and the associated path of the Scout. The region in the blue dotted box corresponds to (C). (\textbf{C}) Online mesh map of the Scout for target identification. (\textbf{D}) Postprocessed high-resolution height map and scientific acquisition paths of the Hybrid and the Scientist. (a and b) Thermal images acquired during the SRC. (c) Example of the rock instance segmentation of a Navcam image. (d) Example of the rock instance segmentation of a CTX image. (\textbf{E}) Example of a panorama image with rock instance segmentation acquired during the SRC. (\textbf{F}) Mission summary of the SRC.}
\end{figure*}

\subsubsection*{Quarry Site, Switzerland}
The quarry site is an active gravel quarry operated by KIBAG, situated in Neuheim, Switzerland. The quarry consists of poorly sorted, fine and coarse sediments, including meter-sized boulders, leading to locomotion challenges such as sinkage and slippage, especially on steep inclines. The site includes a headwall with a maximum slope of about 20$^\circ$. We simulated realistic lighting conditions in a lunar south pole scenario. To this end, we conducted the test at night to minimize the influence of naturally occurring light and illuminated the test site with a 180 W LED lamp (Aputure LS 120D II) at a high illumination angle of roughly 87$^\circ$. As shown in Fig.~4-A, the illumination leads to characteristic long, high-contrast shadows as expected in the vicinity of the lunar south pole. Furthermore, all communication between mission control and the robots passed through a delay simulator, which created a round-trip-time of 5.0 s. \\
We selected distinct boulders on the site to simulate scientific targets of interest. Furthermore, we spread patches of basalt, ilmenite, rutile, and titanium dioxide in different mass fractions on the terrain to create realistic resource-enriched areas for a lunar prospecting mission.

\begin{figure*}
    \centering
    \includegraphics[width=1.0\textwidth]{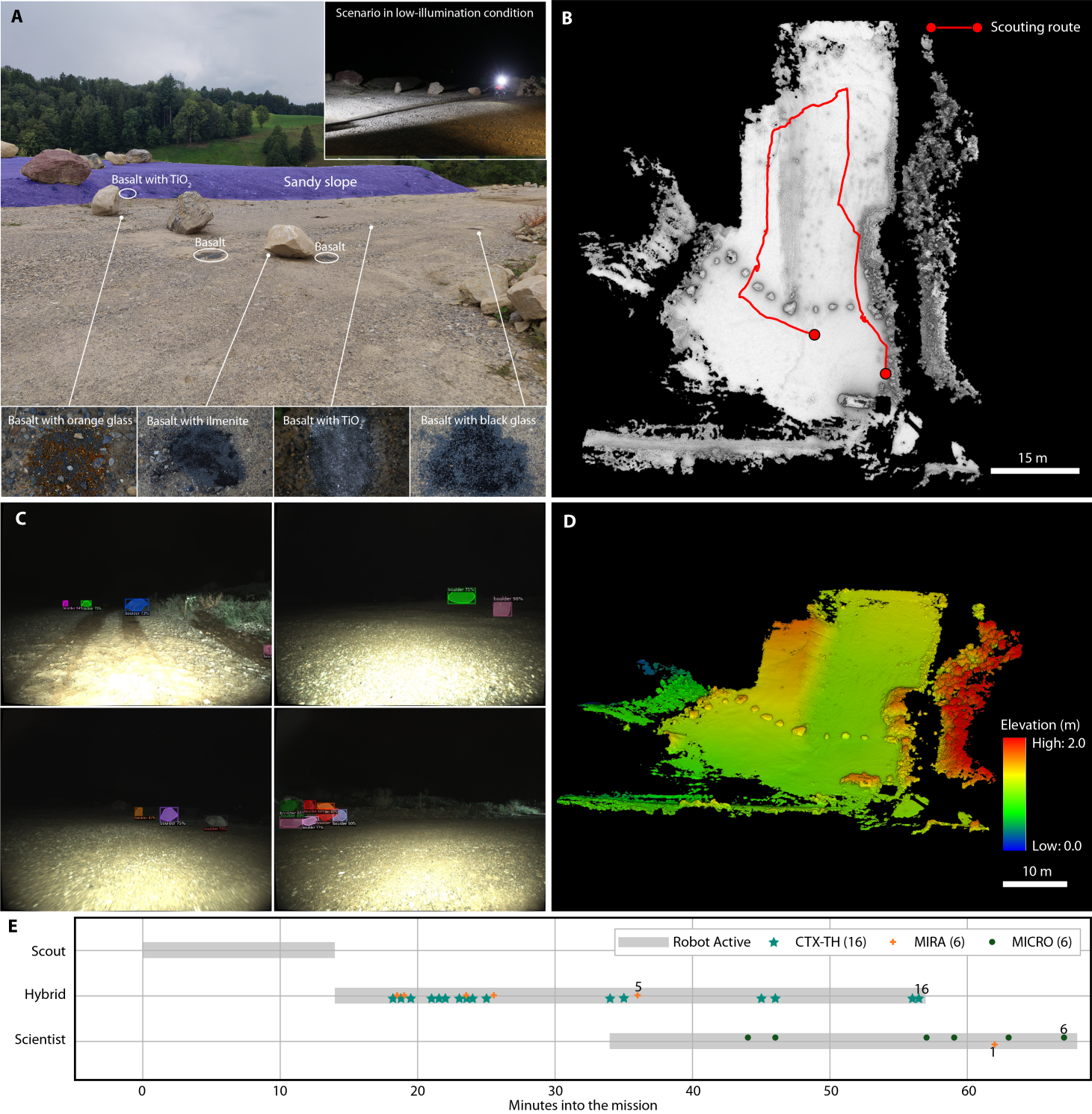}
    \caption{\textbf{Performance of our team of legged robots during the analog mission on the Neuheim quarry site.} (\textbf{A}) Experimental test yard with a variety of scientific targets. (\textbf{B}) Postprocessed and shaded high-resolution map of the Neuheim quarry site with the path of the Scout. (\textbf{C}) Rock segmentation results under low-illumination conditions. (\textbf{D}) Postprocessed high-resolution elevation map of the Neuheim quarry test site. (\textbf{E}) Mission summary of the end-to-end analog deployments at the quarry. The Scout, the Hybrid, and the Scientist were all deployed sequentially with an overlap of the Hybrid and the Scientist. The whole mission duration was 68 min.}
\end{figure*}

\subsubsection*{Locomotion Validation Testbed, Beyond Gravity}
We used the planetary soil testbed at Beyond Gravity (Fig.~S2) to validate the Scout's locomotion controller~\cite{miki2022learning} on steep, granular soil analogs. The testbed was initially designed to test the locomotion subsystem of the ExoMars rover Rosalind Franklin. It features a 6 m x 6 m tiltable container that can be filled with various analog soil simulants. We used ESA's ES-4 martian soil simulant \cite{oravec2021geotechnical} and a row of Jurassic limestone plates as martian bedrock analogs for our tests. The testbed is tiltable up to 25$^\circ$ at a 0.1$^\circ$ resolution.

\subsection*{Analog Mission Results}
This section provides an overview of our analog deployments: the end-to-end deployments at the SRC and in the quarry, as well as the locomotion validation tests.

\subsubsection*{Space Resources Challenge Mission Overview}
During the SRC, the challenge's core objectives provided by the organizers were mapping the competition area, locating boulders and resource-enriched areas (REAs), and characterizing them. Based on that, we derived goals for the robotic system: These comprised of mapping the entire competition area, locating REA candidates and all boulders, and providing scientific data of the boulders and potential REAs to enable trained geologists to characterize them.

Fig.3-F shows the mission overview of the SRC deployment. We first deployed the Scout to map the area and use the navigation cameras and CTX-FW to help the operations team prioritize the targets of interest. After the first LoS, we decided that we had enough targets of interest to deploy the Scientist. The Scientist's task was mainly to focus on potential REAs. After 76 min, we deployed the Hybrid to support the boulder characterization with the thermal imager and the MIRA. 

We mapped 95\% of the competition area, located seven out of eight boulders, and identified 18 potential REAs. We prioritized five boulders for closer investigation, taking CTX-FW images of all of them and MIRA measurements of three boulders. Additionally, we investigated six of the 18 potential REAs with MICRO and MIRA, taking several measurements per target when the data quality was insufficient.

The SRC deployment illustrates several advantages of the teamed exploration approach. Three robots operated simultaneously for 20 min, thanks to the two mission control stations that can be used interchangeably to interact with all robots. During this time, the Scout was completing the map in yet unexplored areas, and the Hybrid and the Scientist were collecting scientific data to characterize the targets of interest. Our autonomy features and intuitive GUI further supported deploying three robots with two operator stations. Additionally, because the environment was already known after the Scout deployment, the Scientist could deploy MICRO and MIRA every 3-5 min, which would be unfeasible with an approach where the same robot has to map and identify targets.

\subsubsection*{Quarry Mission Overview}
To test all payloads on a number of targets with variability, we defined a minimum of five targets to investigate with each instrument. Furthermore, we intended to map a large area to test our localization and mapping systems on a relevant scale. We accordingly defined the following mission goals for the deployment at the quarry: Exploring and mapping an area of at least 1000 m$^2$, and identifying at least five targets of interest, such as boulders or terrain patches. To test our instruments, we set the goal to acquire measurements of at least five targets with each instrument (CTX, filter wheel, thermal imager, MIRA, and MICRO). 

Fig.~4-E shows the mission overview of the quarry deployment. We did not use the CTX-FW on the Scout and limited the use of the MIRA on the Scientist because the hardware malfunctioned. Thus, according to our redundancy concept, we deployed the Hybrid early to take over the CTX tasks of the Scout and the MIRA tasks of the Scientist. The Scientist deployed once the operations team had identified and prioritized enough targets of interest to maximize the payload utilization on the Scientist.

In a total mission time of 68 min, we mapped an area of 1375 m$^2$ and identified twelve targets of interest (ten boulders and two area patches). We collected 16 CTX and thermal images of seven rocks, MIRA spectra of six rocks, and MICRO images of three rocks and two area patches (two MICRO datasets of the same area patch). Using our two control stations, we could efficiently control the Hybrid and the Scientist in parallel during one-third of the mission. Thanks to the redundancy concept in our robotic team approach, we fulfilled all mission objectives except the acquisition of filter wheel images, despite the inactive payloads. A single-robot system or a system without a redundancy concept could not have achieved the objectives related to the CTX and the MIRA measurements under these circumstances.

\begin{figure*}
    \centering
    \includegraphics[width=1.0\textwidth]{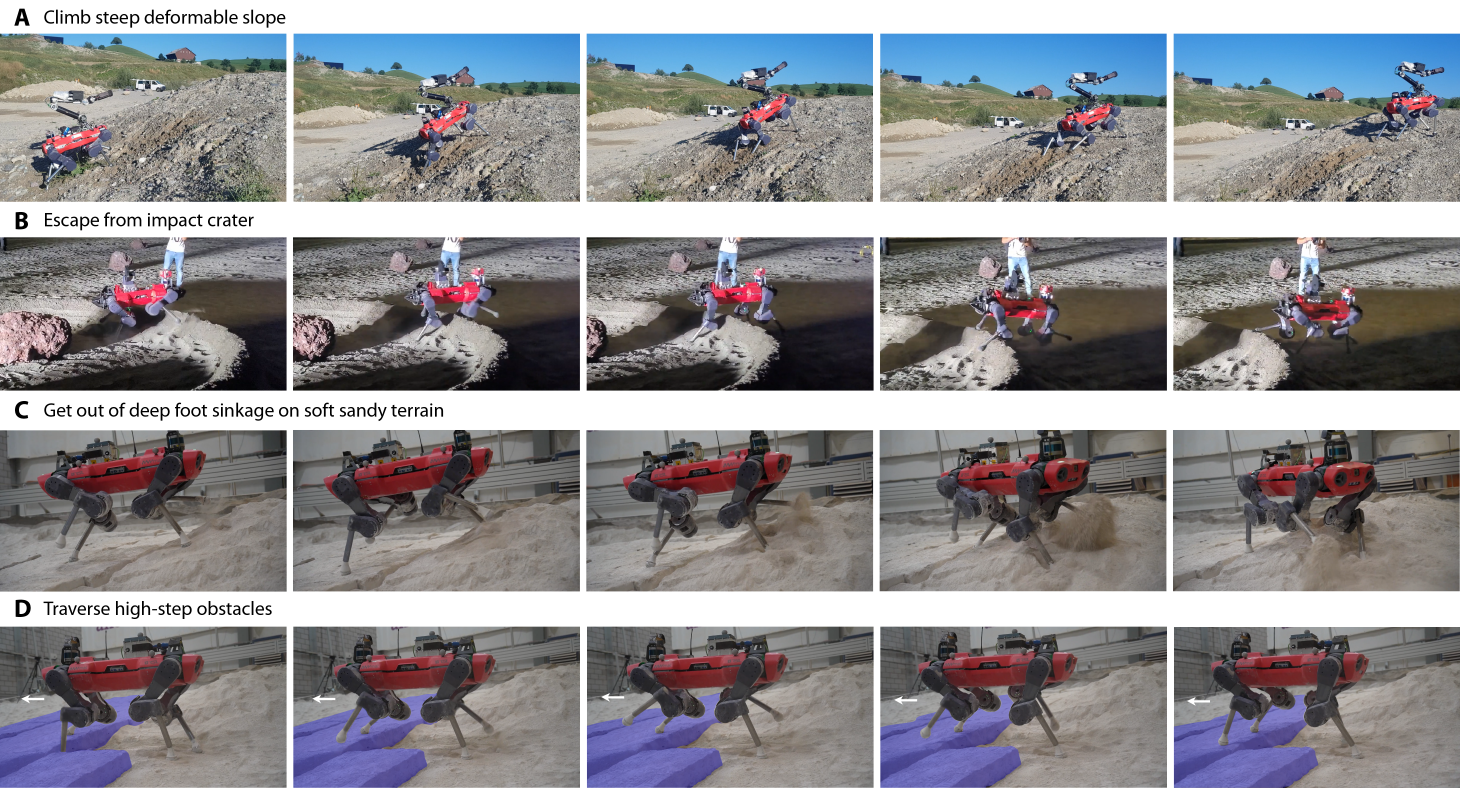}
    \caption{\textbf{Locomotion capability of our customized ANYmal over different challenging planetary analog terrains.} (\textbf{A})  ANYmal climbs a steep sandy slope at around \SI{20}{\degree} on the Neuheim quarry site. (\textbf{B}) ANYmal escapes from the analog lunar impact crater rim during the SRC. \textbf{C}) ANYmal escapes from deep foot sinkage in the locomotion validation test bed. \textbf{D}) ANYmal walks over a high bedrock step.}
\end{figure*}

\subsection*{Locomotion Results}
In the validation test campaign at Beyond Gravity, the Scout could climb and descend slopes of up to 25$^\circ$ - the maximum of the testbed - on ES-4 and bedrock using our existing locomotion control policy \cite{miki2022learning}. On both grounds, we conducted three tests, both in ascent and descent, without a locomotion failure, which we define as either a fall or a failure to continue the traverse, for example due to slippage. Even on the maximum slope of the testbed, the robot reached a top speed of 0.7 m/s, outperforming state-of-the-art systems that can climb such inclinations (Table S3). We conducted additional tests at a 10$^\circ$ slope on ES-4 with steps of bedrock and very loose hills in all directions. The robot could negotiate the hills despite the high sinkage (Fig.~5-C) and seamlessly transition high steps between ES-4 and bedrock (Fig.~5-D).

Our developed controller for the Hybrid and Scientist enables similar robustness as the existing baseline controller~\cite{miki2022learning}, but with added heavy payloads such as the robotic arm. The moving robotic arm did not impede the Scientist while walking on flat terrain (Fig.~S3, Movie~S1). Additionally, with a static arm, the robot consumed 15\% less power in a mock mission on flat ground when using our controller compared to the baseline controller (see Supplementary Results).

During the mission deployments in the quarry and the SRC, the robots traversed steep granular slopes up to 20$^\circ$ (Fig.~5-A), and a crater rim (Fig.~5-B). At the SRC, the robots covered a total distance of 358m of granular terrain (Fig.~3-B and Fig.~3-D). The Scientist, carrying the robotic arm, showed the same level of robustness as the Scout and the Hybrid. During all pre-tests and mission deployments, no single locomotion failure occurred on any of the robots.

\subsection*{Mapping and Target Identification}
We used a lightweight mesh representation for online operations and a dense mapping framework for mission post-processing in both analog missions (Fig.~6). Furthermore, we ran an instance segmentation pipeline on our navigation cameras and CTX images to identify and highlight boulders as potential targets of interest for the operations team, substantially reducing the operational overhead. Fig.~3-C-E and Fig.~4-C-E show the mapping and perception results in the quarry and at the SRC, respectively.

The robot sent a downsampled point cloud in a nine-meter radius around the robot to mission control. The mission control PC generated the mesh and fused single mesh instances automatically to gradually build a mesh map of the covered area. Offloading the meshing operation to the mission control PC allowed us to transmit a small point cloud instead of a heavy map. Fig.~3-C shows a fraction of the mesh map of the SRC built with three mesh instances. Boulders 4 and 5 and impact craters 4 and 5 are distinguishable in the mesh. Together with the navigation cameras, the resolution of the mesh map allowed the operations team to identify and mark these targets of interest for further investigation. Therefore, the resolution was high enough to understand the robot's environment, select targets on the mesh map, and make mission scheduling decisions while still allowing the map to be transferred over the network.

Fig.~3-D-{c,d} and Fig.~4-C show the instance segmentation output on the navigation cameras in the SRC and quarry missions, respectively. The images show that, even under difficult lighting conditions, the pipeline can identify and highlight boulders in the image, supporting the operations team in the target identification. Additionally, we segmented selected panoramic images (Fig.~3-E). 

Aside from the online mesh to guide operations, all robots maintained a high-resolution point cloud map in both missions for post-mission visualization. Fig.~3-B shows the post-processed and shaded dense map of the complete SRC competition environment, with the path of the Scout overlaid. We explored 95\% of the area with the Scout within 96 min, where most of the environment and the scientific targets were already identified after 50 min. Similarly, Fig.~4-B shows the dense map created during the quarry mission. Additionally, we colored the dense maps by elevation (Fig.~3-D and Fig.~4-D) to create a topographic overview of the environment. For example, the headwall in the quarry is visible in the top left part of the map. Furthermore, we created colored maps (Fig.~S4 and Fig.~S5) by projecting the navigation camera colors onto the point cloud.

\begin{figure*}
    \centering
    \includegraphics[width=1.0\textwidth]{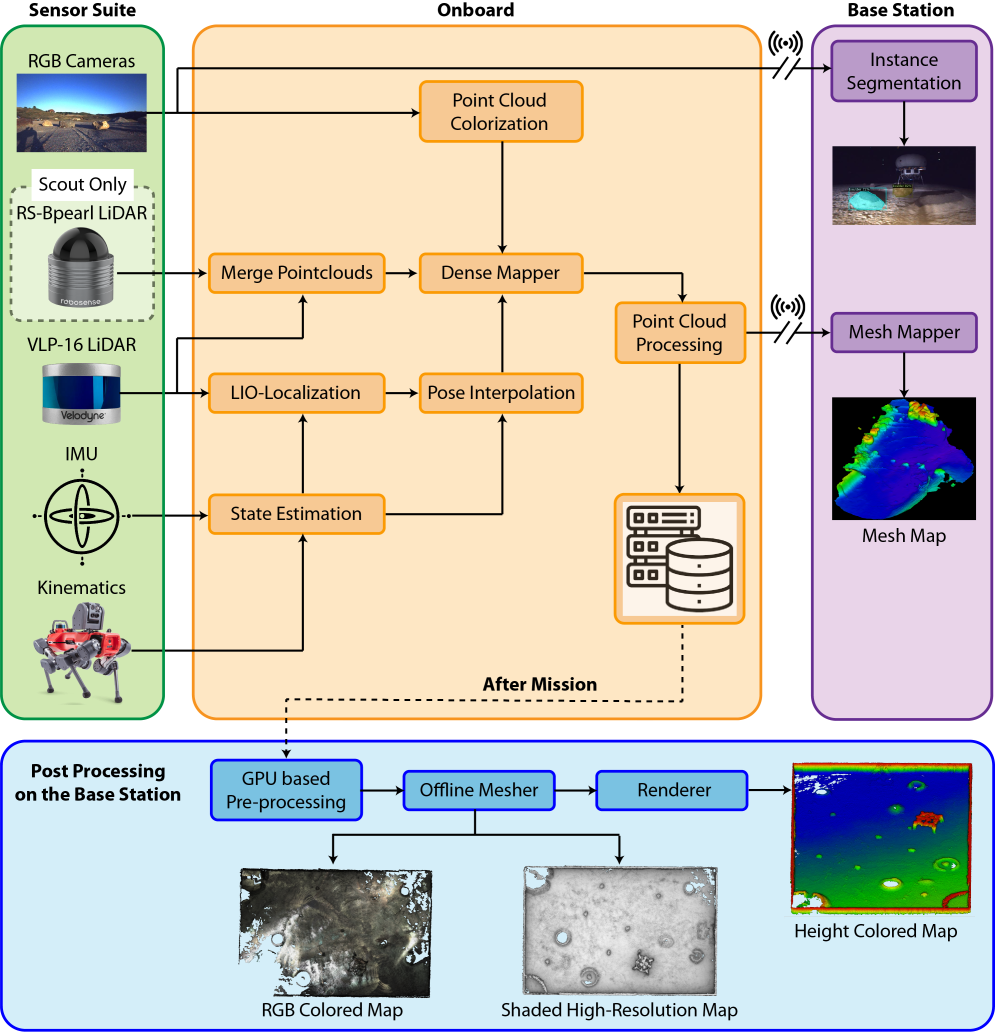}
    \caption{\textbf{ Mission support mapping and localization modules.} The blue area represents the postmission operations, whereas the other modules are active during the mission.}
\end{figure*}

\subsection*{Scientific Data of Targets of Interest}
Once the operations team identified and prioritized targets of interest, they could send remote measurement and in-situ measurement tasks to all robots (Fig.~7). Fig.~8 shows an example of the scientific data we could gather during the SRC. We acquired images in five visual spectral ranges at two zoom levels using CTX-FW (Fig.~8-A). The low-zoom image shows the target's size, shape, and geomorphic context. The high-zoom image provides information on the target's surface texture, including the presence of millimeter to centimeter-scale vesicles, the rock's lithology, and the distribution of minerals. Both low-zoom and high-zoom images indicate that all boulders in the SRC were porous basalts with aphanitic (fine-grained) textures. The high-zoom images with the filter wheel enabled us to study the relative reflectance of each target and compute five-point spectra. We color-corrected selected RBG images with images from a robot-mounted color calibration card.

\begin{figure*}
    \centering
    \includegraphics[width=1.0\textwidth]{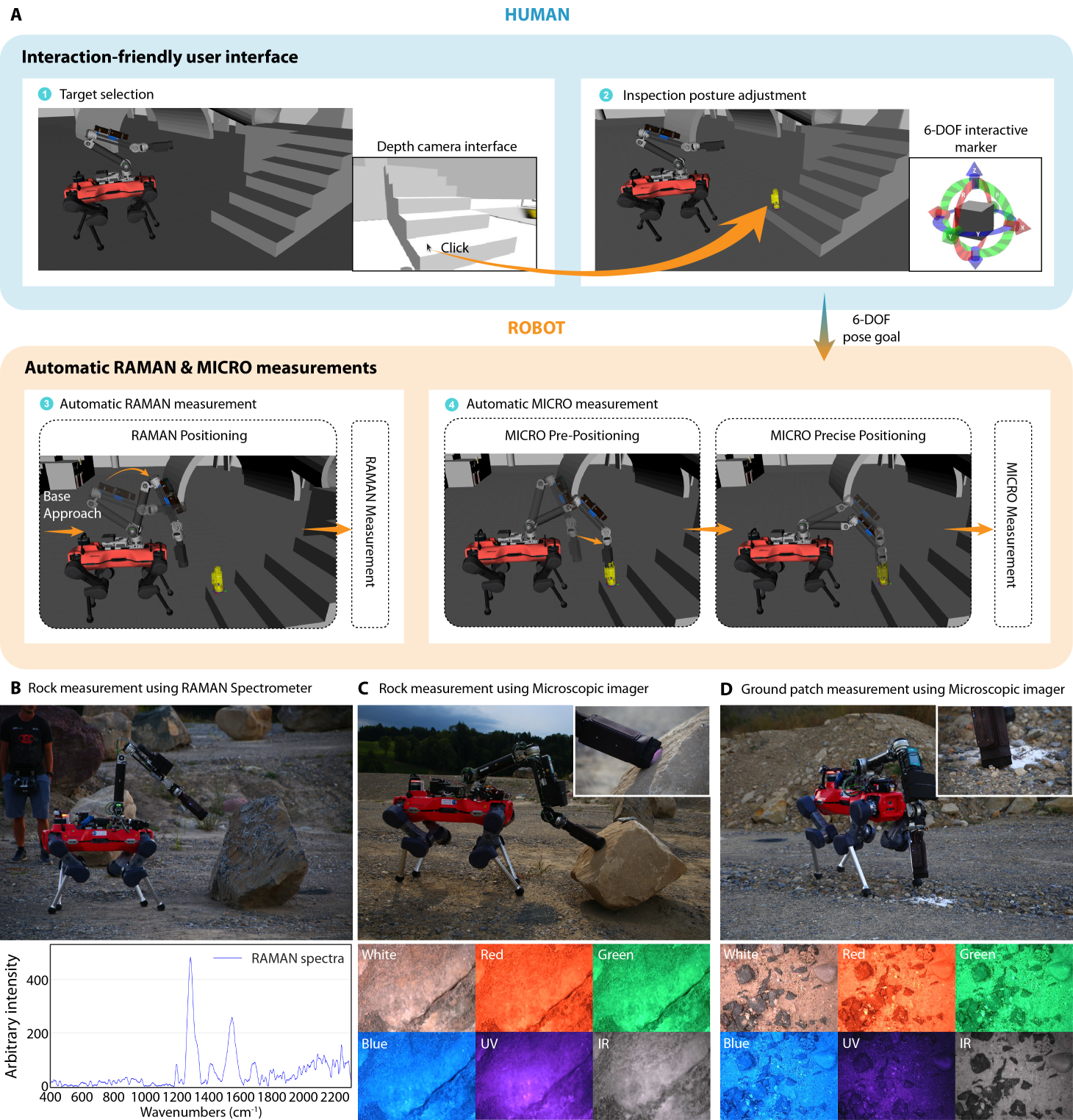}
    \caption{\textbf{In situ measurement task workflow.} (\textbf{A}) The operator selects the desired target in the depth camera interface to define the instrument pose. The pose can be adjusted using a 6-DoF interactive marker. The robot receives the 6-DoF goal and uses it to deploy the scientific instruments at the desired location. (\textbf{B}) Boulder measurement using the MIRA Raman spectrometer and associated data products. (\textbf{C})  Boulder measurement using MICRO and associated data products. (\textbf{D}) Ground patch measurement using MICRO and associated data products.}
\end{figure*}

Fig.~8-B shows in-situ measurement results of a potentially resource-enriched area during the SRC. We collected microscopic images in six different spectral bands (white, red, green, blue, UV, IR), allowing us to compute the target's five-point spectrum. The images show the coarse granular basalt regolith, which covered the whole competition area at the SRC. MICRO images enabled us to investigate the grain size distribution and presence of potential resources. The MIRA Raman spectrum of the basaltic regolith in Fig.~8-B shows, for example, a prominent peak at 952 cm$^-1$.

\begin{figure*}
    \centering
    \includegraphics[width=1.0\textwidth]{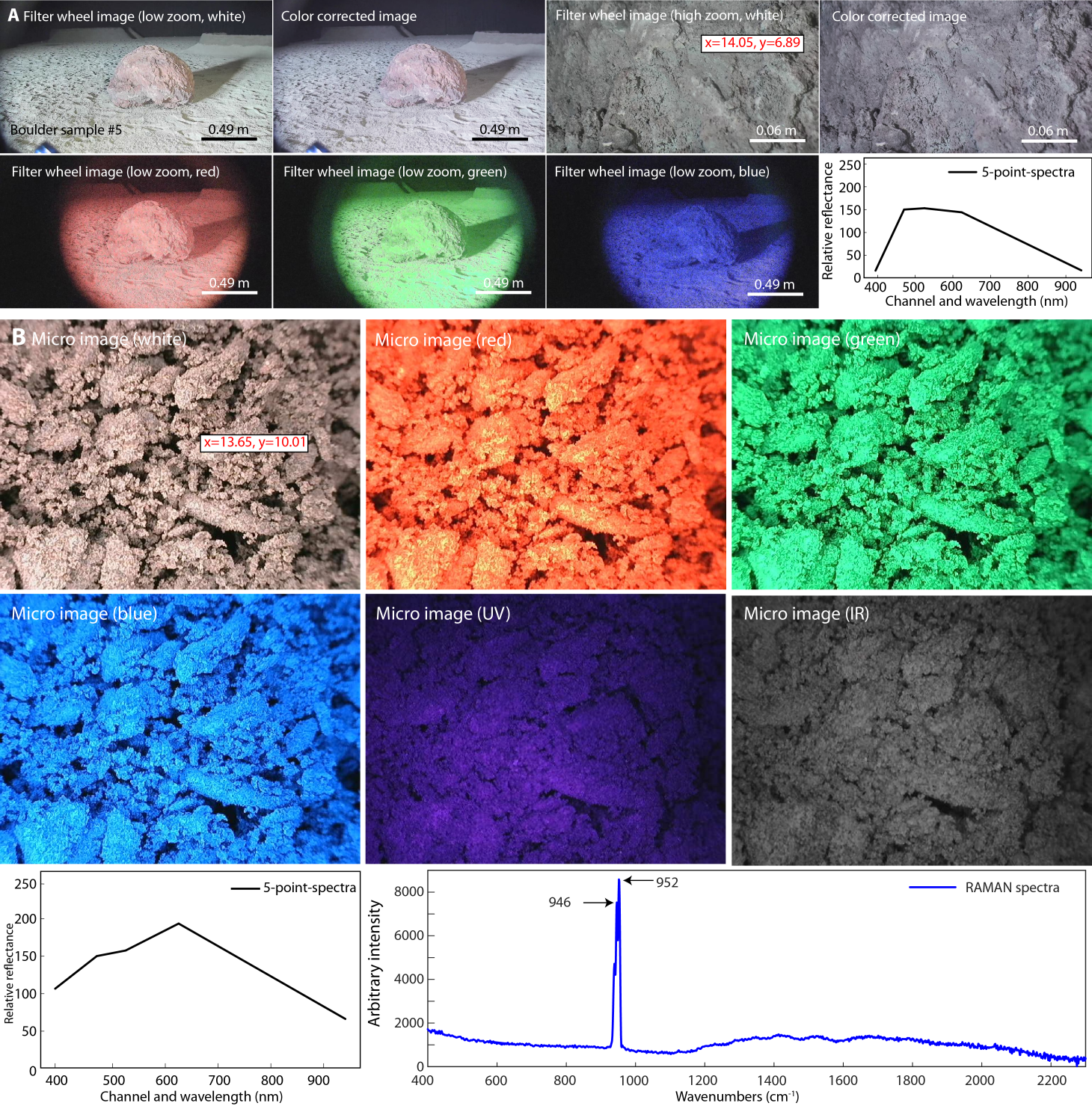}
    \caption{\textbf{Examples of scientific data products at the SRC.} (\textbf{A})  Multispectral visual data acquired by the CTX-FW of boulder 5. The images show a vesicular basaltic boulder. (\textbf{B}) MICRO and Raman data products of a REA candidate. The images and spectra show the basaltic soil of the competition area. The relative reflectance has an arbitrary scale.}
\end{figure*}

\section*{Discussion}
Using a team of legged robots with powerful locomotion capabilities, a mapping pipeline for online and offline visualization, and segmentation tools for target identification allowed us to collect a substantial amount of scientific data in planetary analog missions with limited mission time. Compared to the single-robot approach we deployed at the first Space Resources Challenge field trial~\cite{arm2022results}, we could drastically increase the quality and quantity of scientific data products. The specialization of our robots allowed a high payload utilization, as proven at the SRC, where the Scientist robot deployed the MIRA and MICRO instruments every 3-5 min. In a multi-robot approach without specialization, for example without a dedicated scout robot, every robot would invest a substantial part of the mission time in mapping and target identification, limiting the high-return instruments' payload utilization.

The quarry deployment showed the importance of the redundancy concept. We could still fulfill six out of seven mission goals despite two malfunctioning payloads. Although we did not experience a robot failure in the deployments, our redundancy concept would allow us to still accomplish most mission goals.

Our payload selection features a balanced mix of remote and in-situ science, but it was subject to budget and time constraints. Other instruments, for example X-ray fluorescence spectrometers (XRF), as used by other teams at the SRC, performed better at identifying REAs. Further expanding our scientific instrument suite, for example, with an XRF or laser-induced breakdown spectrometer (LIBS), would increase the quality and diversity of the mission's science outputs. Additionally, adapting our system to allow sample return to a lander with more capable scientific instruments would be a valuable expansion to increase the mission's scientific output.

One learning from our field deployments is that a round-trip-time of 5s in the communication between mission control and the robot notably affects operations. Standard, reliable protocols such as TCP are not suitable in this setting. When using UDP, we had to reduce the data size of products, such as point clouds and images, to decrease the probability of a single packet loss occurring within the transfer of a data product. Furthermore, every operator interaction costs valuable mission time. We tackled this issue by providing lightweight, expressive data to the operators, such as the mesh map, and employing task-level autonomy. Hence, every task execution only requires one operator interaction. However, a more in-depth analysis of suitable protocols and compression technologies could further alleviate this problem. 

A key learning in our deployments was the importance of the human aspect: The robots could execute single tasks autonomously, but the human operators still had to prioritize and allocate tasks. This decision-making can be time-consuming and difficult, especially since the operators must balance the time required for the decision-making process with the mission time. An algorithmic approach to decision-making and task allocation could improve the mission return. The algorithm could either aid the operations team or, ideally, be implemented in an autonomy module on the robots to allow longer-term or even full mission autonomy. However, besides semantic scene understanding, this requires automatic interpretation of scientific data products such as images and spectra, which is an open challenge. Furthermore, more robot-to-robot communication is needed to build collaborative maps, allocate tasks without mission control in the loop and allow for more involved interaction and collaboration between multiple robots.

A higher level of autonomy will additionally improve the system's scalability to applications with even more challenging communication, such as Mars exploration. Moreover, it will allow scaling the approach to a higher number of robots without increasing the workload on the operations team. Required modules to increase the level of autonomy are a safe, multi-goal planner to visit identified targets, automatic target identification and prioritization as stated above, and automatic task allocation to the robots based on the robot-specific skillset. 

In this work, we used legged robots with different scientific investigation skills but almost identical locomotion skills. In future deployments, a more heterogeneous approach could be advantageous. Drones would enable fast terrain mapping, and wheeled robots, with limited mobility but higher battery life, could investigate easy-to-reach targets while the legged robots focus on hard-to-access areas.

Lastly, a relevant research stream is the scalability of such prototypes from analog deployments to actual space missions~\cite{kolvenbach2018scalability}. Challenges specific to legged robots mainly concern the power concept and thermal management. Preliminary results show that these aspects are feasible~\cite{kolvenbach2018scalability,valsecchi2022preliminary}. However, there are currently no powerful space-graded processors to run the mapping, navigation, and locomotion algorithms in real-time. Additionally, our localization and mapping pipelines heavily rely on LiDARs, which are currently not space-qualified. However, solid-state LiDARs could pave the way for LiDAR-based localization and mapping in space applications.

Although there remains substantial room for future research and several open challenges in scaling the presented system to actual space missions, this work is an important step towards teams of legged robots for planetary exploration that provide value in both scientific and commercial missions.

\section*{Materials and Methods}

\subsection*{Hardware Description}
All described robots in this report are based on ANYmal by ANYbotics. The Scout is an ANYmal C (2019), whereas the Hybrid and the Scientist are based on ANYmal D (2021). Additionally, the Scientist carries a DynaArm, a custom 6DoF robotic arm. ANYmal weighs 50 kg, has a payload capacity of 15 kg, and a nominal operation time of 90 min while continuously walking. We show an overview of the on-board computers in Table~S4. 

\subsubsection*{Robotic Payloads}
In the factory configuration, the ANYmals are equipped with a VLP16 Puck LITE LiDAR by Velodyne for mapping and localization, four (ANYmal C) or six (ANYmal D) RealSense D435 active stereo sensors by Intel (front, left, right, rear) for elevation mapping (Sec. Elevation Mapping), and two FLIR Blackfly wide-angle cameras (front and rear) to provide RGB image streams. We customized all robots with two high-power, air-cooled LEDs at the front and rear to illuminate dark environments.
Since the Scout is the desired mapping robot, we exchanged the four RealSense D435 with two Robosense RS-Bpearl LiDARs (front and rear) for more accurate elevation mapping (Sec. Elevation Mapping) and dense point cloud mapping (Sec.~Dense Point Cloud Mapping). Additionally, we added Sevensense's Alphasense Core visual-inertial sensor as a high-performance navigation camera array on the Scout. It features three monochrome and four color cameras (front, left, right, and top). Each camera has 0.4 MP and a field of view of 126$^\circ$ x 92.4$^\circ$ (HxV). In addition, four high-power, air-cooled LEDs are encapsulated in the same housing. The Alphasense Core allows the operator to quickly gain a first overview in an unknown environment and enables RGB-colored mapping (Sec.~Dense Point Cloud Mapping).

\subsubsection*{Science Payloads}
The Scout carries an ANYbotics inspection payload, comprising a 10x optical zoom camera and a spotlight on a pan-tilt unit. We augmented the zoom camera with a spectral filter wheel. The filter wheel contains five narrow band-pass filters (390 nm, 470 nm, 530 nm, 620 nm, and 940 nm) and one position with no filter. In this configuration, we refer to the payload as CTX-FW. We selected CTX-FW to obtain high-resolution images of the targets and their surroundings as well as multi-spectral information since lunar minerals are known to exhibit distinct reflective features at different wavelengths. On the back of the robot, we mounted a color calibration target with 24 color squares to color-calibrate CTX images.

The Hybrid carries a newer ANYbotics inspection payload for the ANYmal D. It consists of a 20x optical zoom camera, a thermal camera, and a spotlight on a pan-tilt unit. We refer to this payload as CTX-TH. Aside from high-resolution images, CTX-TH enables the thermophysical analysis of targets. Rocks and regolith have different thermal signatures depending on their composition. Additionally, we equipped the Hybrid with a Metrohm Instant Raman Analyzer XTR DS (MIRA), a ruggedized, portable Raman spectrometer~\cite{mira}. MIRA is equipped with an auto-focus attachment that enables measurements of targets from a distance of up to two meters. The MIRA allows compositional analysis of the target's mineralogy, which provides a more in-depth investigation than the CTX-FW and CTX-TH imaging techniques.

The Scientist uses the DynaArm to deploy the instruments for in-situ measurements. The DynaArm has a total weight of 10 kg, including the 2.3 kg of the scientific payloads. It has a reach of 0.9 m without any tools attached and a nominal payload capacity of 8 kg at half of the reach with a maximum tool speed of up to 13.5 m/s. The forearm of the DynaArm carries a MIRA. Additionally, we mounted our custom-built UV-VIS-NIR microscopic imager (MICRO) as the end-effector (Fig.~S1-B). It contains a USB microscope (Dinolite AD4113T-I2V) mounted on a linear actuator mechanism, a ring of 48 RGB LEDs, a time-of-flight sensor, and control electronics. The front of the casing is equipped with a foam ring, which ensures that no stray light enters the case when MICRO is in contact with a target. The internal UV and NIR LEDs of the Dino-Lite microscope and the RGB LED ring allow for acquiring multi-spectral microscopic images in the entire spectral range from UV to NIR at 395 nm, 470 nm, 525 nm, 620 nm, and 940 nm. The time-of-flight sensor determines the distance between the microscope and the target. The distance feedback allows the microscope, in combination with the linear actuator and an autofocus routine, to acquire sharp images of the samples, independent of the precise placement of the instrument. Using MICRO's multi-spectral images, trained geologists can perform petrographic assessments which are more detailed than the analyses based on CTX-FW.

\subsection*{Legged Locomotion in Planetary Environments}
We use a reinforcement learning (RL) approach to design our locomotion policy because RL approaches have shown robust performance in challenging environments \cite{lee2020learning,peng2020learning,miki2022learning}. We base our work on the perceptive locomotion pipeline originally published in \cite{miki2022learning}, which has already been successfully used in other field deployments such as the DARPA SubT challenge \cite{tranzatto2022team}. 
In this pipeline, the control policy is trained in two stages: First, we train a teacher policy with all ground truth information. Then, we train a student policy to output the same action as the teacher from noisy and limited information.

We use the existing controller from \cite{miki2022learning} on the Scout. On the Hybrid and the Scientist, however, the scientific payloads and the robotic arm can lead to big disturbances for the controller,  resulting in reduced robustness and inefficient motion. Therefore, we developed an updated locomotion policy with explicit payload measurements: We add payload mass randomization during the training and explicitly give the mass information to the teacher policy. Secondly, we add arm position and velocity observations to allow the policy to actively counteract the wrench on the base caused by the arm's motion. Additionally, we increased the joint torque reward to further improve energy efficiency.
During training, we gave a random arm initial position and random arm target position and controlled each arm joint via PD control to simulate the arm movements. We present more detail on our locomotion setup in the Supplementary Methods.

\subsection*{Localization, Mapping, and Perception}
In this section, we describe our localization and mapping system, designed to globally localize the robots and generate an accurate representation of the environment during the mission. Furthermore, we describe the instance segmentation algorithm that helps the operations team understand the environment.

\subsubsection*{Robust Lidar-Based Localization}
We rely on LiDAR-based SLAM because of the long range and large field of view of LiDARs. A core challenge of LiDAR-based SLAM in planetary analog missions is LiDAR degeneracy. To cope with degenerate environments, we use a modified version of CompSLAM~\cite{compSLAM}, which has shown to perform well under degeneracy through numerous deployments~\cite{tranzatto2022cerberus}. CompSLAM is a complementary multi-modal SLAM system based on the Iterative Closest Point algorithm~\cite{p2planeICP}. CompSLAM can use visual, thermal, or inertial data as a robust prior of the LiDAR mapping module. In this work, we rely only on the inertial data as a prior. If the environment degenerates, CompSLAM identifies degenerate directions and directly uses the prior to integrate the point cloud scan into the map. 

\subsubsection*{Dense Point Cloud Mapping}
CompSLAM maintains only a sparse map to keep the SLAM optimization problem tractable. We therefore require a separate module for dense mapping to create a high-resolution environment representation. To this end, we maintain an octree-based map with a voxel size of 30 mm on the robots.

Firstly, we merge and filter the point clouds of all LiDAR sensors on the robot. After registering the point cloud into the dense map using the pose estimate of the SLAM system, we clip the map at a two-meter height to remove unnecessary data. Furthermore, we apply an outlier filter by removing points that do not have at least five neighbors within a 200 mm radius. Lastly, we apply a multi-robot crop filter to reject points that lie on the other robots. The mapping pipeline additionally maintains a colorized map by projecting the RGB information from the navigation cameras onto a copy of the point cloud.

After the mission, we fetch the high-resolution point cloud map from the robot, simulating long-term data transmission. First, a GPU-based pre-processor calculates the surface normals at each point using the nearest 40 points. Subsequently, we use a statistical noise filter and a radius filter to filter the map. The statistical noise filter considers the nearest 10 points and filters out-of-distribution points (standard deviation $> 2 \sigma$). After the pre-processing, we reconstruct a triangle mesh with a tree depth of 12 using the Poisson surface reconstruction method \cite{kazhdan2006poisson} included in the open3D library \cite{open3d} to generate the high-resolution, continuous mesh representation. This process takes up to five minutes on a computer with an Intel i9 12700K CPU depending on the terrain complexity. Later, the generated high-resolution mesh is visualized with an eye dome lighting shader \cite{edl}.

\subsubsection*{Lightweight Mesh Representation}
Since an unreliable, high-latency network cannot transport the dense map in real-time, we periodically send a downsampled version of the dense map (voxel size of 150 mm) in a 9 m radius around the robot to mission control. Before transmission, we compress the point cloud using the Draco point cloud compression library~\cite{draco}. The mission control PC uses this downsampled and compressed point cloud to create a lightweight mesh representation~(Fig.~S6). With downsampling and compression, we reduce the size of the point cloud that is transported over the high-latency network from 3 MB to 250 kB. Furthermore, the robot only sends the point cloud if it contains substantial new information, for example when the robot moves, or when the user requests a mesh of a certain area in the user interface. The meshing operation on the mission control PC is limited to a maximum of 0.25 Hz.

The mission control PC calculates a triangle mesh using the Poisson surface reconstruction method \cite{kazhdan2006poisson} with a depth parameter of eight. We recalculate the surface normals of the mesh to correct the orientation and smoothness of the mesh surfaces. Additionally, we filter out vertices and triangles with less than ten support points. This mesh is then merged into the mesh map by fusing the non-overlapped vertices and triangles and averaging the overlapping areas.

\subsubsection*{Elevation Mapping}
The elevation map is a local robot-centric 2.5D grid representation where each cell indicates the terrain height. We use it as a terrain representation with a high update rate for the local planning module (Sec.~Autonomous Local Navigation). The elevation map is 8 m x 8 m with a resolution of 40 mm. We use an elevation mapping pipeline running on a GPU \cite{miki2022elevation} to integrate the point cloud data into the elevation map in parallel, based on the robot's odometry.

\subsubsection*{Instance Segmentation}
We apply a boulder instance segmentation network to autonomously identify each boulder instance in the RGB images to contextualize the scientific data. We built our approach upon Mask R-CNN \cite{he2017mask}. The network predicts each boulder instance's bounding box, outline, and confidence. We finetuned the original model on a custom-built dataset containing hundreds of images collected in the first field trial of the SRC with instance labels. In this custom-built dataset, ANYmal acquired the images in a dark analog lunar terrain, similare to the setting in the final challenge. We deployed our rock instance segmentation network on the mission control PC to segment navigation and CTX images. Fig.~S7 shows several prediction results, indicating the network's robust performance under variable light conditions and different input cameras. 

\subsection*{Scientific Payload Integration and Deployment}
The operator can send 3D remote measurement targets to request images from CTX-FW or CTX-TH. For CTX-FW, the operator can additionally specify which spectral filters to use. The operator marks a target on the map or uses an already-marked target to send a remote measurement task. When requesting RGB images, the operator can select between a high and a low zoom level, with an image width of 0.3 m and 3 m in the focal plane, respectively. The robot sends the images to mission control, which automatically saves them by the target identifier. Additionally, the operator can request panoramic images. The robot then takes nine images at a fixed zoom level, which are stitched in post-processing.

The MIRA communicates to the onboard PC of the robot through a USB interface. It is fully integrated into the software stack through a custom ROS wrapper that provides a ROS action interface to trigger a measurement. The measurement procedure is then fully automated: the auto-focus attachment measures the distance to the target, focuses on this distance, and then the MIRA initiates a measurement. The MIRA matches the received spectroscopic response against the custom-built library of 355 different spectra of relevant lunar minerals (for example oxides, olivines, pyroxenes, and feldspars) (Table~S5). The robot sends the spectroscopic response and the corresponding match to mission control for the operations team to evaluate.

The MICRO has a USB interface and communicates to the onboard PC of the robot through rosserial. As for the MIRA, its measurement sequence is fully automated. On a measurement request, the device focuses on the target automatically, records an image sequence at all available spectral bands (UV, R, G, B, NIR), and returns a complete measurement data package.

\subsubsection*{Payload Deployment with the Robotic Arm}
To deploy the MIRA, the arm aligns the Raman laser with the target at the desired distance of 0.7 m. The deployment of MICRO happens in two stages: Firstly, the arm aligns the MICRO with the target pose at a predefined offset of 100 mm in the inverse heading direction of the target pose. In the second stage, a PID controller moves MICRO along the heading axis of the target pose to bring the front foam piece (Fig.~S1) in contact with the target surface using the ToF distance measurement as feedback. The two-stage approach allows MICRO to accurately make contact with the target surface, even if the original 6D in-situ measurement target is not precisely on the surface. We provide information on the arm controller in the Supplementary Methods.

\subsection*{Autonomy}
We use behavior trees (BT) \cite{faconti2019mood2be} to handle the autonomous task execution on the robots. The modularity of BTs allows us to reuse modules across the different robots. The BT on the robot processes all operator interactions and starts, monitors, and stops the requested task. In this way, the operator sets high-level objectives while retaining the ability to stop or change the task easily. We show more details of our behavior trees in the Supplementary Methods.

\subsubsection*{Autonomous Local Navigation}
The operators guide the robots via high-level navigation goals. We specifically avoided a fully autonomous approach because we preferred the operations team to select and prioritize scientific targets during the mission in unknown environments.

To allow the operator to safely operate the robots via waypoints, we use a sampling-based local planner \cite{wellhausen2021rough} that operates on the local elevation map. Unlike most state-of-the-art navigation planners, it does not assign traversability values to discrete terrain patches. Instead, the robot morphology is approximated by reachability volumes for the feet and a collision volume for the torso, which makes the planner more suitable for legged robots. Using a PID controller, a path-tracking module converts the traversable path to twist commands for the locomotion controller.

\subsubsection*{Autonomous In-Situ Measurement Acquisition}
Once the Scientist is close to a target, the operator can select the in-situ measurement point on the grey-scale infrared image stream of the Realsense camera. We use the grey-scale infrared image from the Realsense camera because it does not provide an RGB image stream and the infrared image provides good output in bad lighting conditions. The user interface spawns a 6-DoF interactive marker, which can be adjusted if necessary. The operator can specify the exact measurement task, that is, a MIRA measurement, a MICRO measurement, or both.

Once the robot receives the measurement task, it approaches the target and aligns the base at a predefined offset of 1 m to the target. Then, the robot deploys the instruments and triggers a measurement automatically. Depending on the task definition, the robot takes a MIRA measurement, a MICRO measurement, or both. If an error occurs during the measurement process, the arm will move back to the default position, and the robot will notify the operator.

\subsubsection*{Loss of Signal Operation}
If a LoS occurs, the robots finish their currently allocated task and then switch to a dedicated LoS behavior. The Scout and the Hybrid perform an autonomous measurement routine. First, they record a panoramic image of their environment. Second, they image their own footprints, which provide information about soil mechanics. Third, the robots acquire images of each other and the Scientist to assess the status of the hardware. Fourth, the Scout takes an image of the color calibration card. This image is used in post-processing to perform color correction. The robots send the acquired data to mission control once communication is restored. The Scientist remains in a nominal standing configuration since the system currently does not autonomously detect new measurement targets.

\subsection*{Resilient Communication for High-Latency Networks}
Resilient communication in high-latency networks is crucial to ensure data transmission between mission control and the robots. We use commercial off-the-shelf radio devices by Rajant \cite{rajant} to create a reliable mesh network. The mesh setup includes the base station and the robots, each acting as a mesh node. The robots are equipped with a BreadCrumb DX2 and the base station consists of a BreadCrumb ES1 with a panel antenna. All radios operate at 5.8 GHz. The base station and the mission control PCs are connected by wire via a delay simulator that ensures a 5.0 s RTT delay.

We use the Robot Operating System (ROS1) \cite{quigley2009ros} on all robots and mission control stations to manage onboard communications. Each robot and mission control station had a separate rosmaster. However, ROS1 is not designed to communicate over high latency networks, as a TCP handshake is required to establish a connection, even if the data is transmitted over UDP. Therefore, we use Nimbro Network to establish communication between different rosmasters \cite{stuckler2016nimbro,schwarz2017nimbro}. Nimbro Network was specifically designed to work in unreliable high-latency networks and used in the SpaceBot cup, a robot competition for lunar exploration \cite{kaupisch2015dlr}.

Using the predominant Internet protocol TCP is problematic because of the RTT delay and high bandwidth. Consequently, we transmit all data via UDP. UDP does not provide congestion control, and there is no guarantee that data will arrive. Therefore, mission control and the robots only exchange essential data. We show the network usage of mission control 1 at the SRC in Fig.~S8. The robot-to-robot communication, however, runs on TCP because the delay is in the low millisecond range.

\textbf{Acknowledgments:} We would like to thank the implementation partners, namely the Lucerne University of Applied Sciences and Arts (HSLU), ANYbotics AG, and the maxon SpaceLab. A special thanks to Samuel Tenisch, Alex Brandes, Gerhard Székely, and Nico Steinert for developing the filter wheel, Fabio Mast, Lars Horvath, Hugo Umbers, and Marco Trentini for supporting the MICRO development, and Daria Larcher and Julian Bernasconi for supporting the scientific data analysis. We would like to thank Metrohm Schweiz AG for providing the Raman spectrometer. We acknowledge the ETH Earth Sciences Collections and the University of Basel for supporting and providing rock and mineral samples for our experiments. We thank KIBAG Kies Neuheim AG for making the quarry available. This work has been conducted as part of ANYmal Research, a community to advance legged robotics. \textbf{Funding:} This research was supported by the Swiss National Science Foundation (SNF) through the National Centre of Competence in Research Robotics (NCCR Robotics), through the National Centre of Competence in Digital Fabrication (NCCR dfab), and by an ETH Zurich Research Grant No. 21-1 ETH-27. This project has received funding from the European Research Council (ERC) under the European Union’s Horizon 2020 research and innovation programme grant agreements No 852044 and No 101016970. This project has received funding through ESA Contract No. 4000137333/22/NL/AT and 4000135310/21/NL/PA/pt. \textbf{Author contributions} All authors contributed to the system design and wrote the paper. PA, GW, JP, TT, RZ, VB, GL, FK and HK participated in the field deployments and evaluated the respective data. PA, GW, JP and TM conducted the additional locomotion tests. TT and RZ developed the mapping and perception pipelines. JP developed the arm controller. TM developed the locomotion controller. VB, GL and FK selected and designed the scientific payloads and evaluated the respective data. \textbf{Competing interests: } The authors declare no competing interest. \textbf{Data and materials availability:} All data needed to evaluate the conclusions in the paper are present in the paper or the Supplementary Material. Other material is available at Zenodo (DOI: 10.5281/zenodo.8019960).

\section*{Supplementary materials}
\makebox[1.8cm][l]{Supplementary Results} \\
\makebox[1.8cm][l]{Supplementary Methods} \\
\makebox[1.8cm][l]{Tables S1 to S6} \\
\makebox[1.8cm][l]{Figures S1 to S13} \\

\bibliography{scibib}

\begin{thebibliography}{10}

\bibitem{qiao2021ina}
L.~Qiao, J.~W. Head, L.~Wilson, Z.~Ling, Ina lunar irregular mare patch mission
  concepts: distinguishing between ancient and modern volcanism models, {\it
  The Planetary Science Journal\/} p.~66 (2021).

\bibitem{glotch2021scientific}
T.~D. Glotch, E.~R. Jawin, B.~T. Greenhagen, J.~T. Cahill, D.~J. Lawrence,
  R.~N. Watkins, D.~P. Moriarty, N.~Kumari, S.~Li, P.~G. Lucey, M.~A. Siegler,
  J.~Feng, L.~Breitenfeld, C.~C. Allen, H.~Nekvasil, D.~A. Paige, The
  scientific value of a sustained exploration program at the aristarchus
  plateau, {\it The Planetary Science Journal\/} p. 136 (2021).

\bibitem{smith2020artemis}
M.~Smith, D.~Craig, N.~Herrmann, E.~Mahoney, J.~Krezel, N.~McIntyre,
  K.~Goodliff, The artemis program: an overview of nasa's activities to return
  humans to the moon, {\it 2020 IEEE Aerospace Conference\/},  1--10 (IEEE,
  2020).

\bibitem{colaprete2021volatiles}
A.~Colaprete, Volatiles investigating polar exploration rover (viper), {\it
  NASA Technical Reports\/}  (2021).

\bibitem{colaprete2010detection}
A.~Colaprete, P.~Schultz, J.~Heldmann, D.~Wooden, M.~Shirley, K.~Ennico,
  B.~Hermalyn, W.~Marshall, A.~Ricco, R.~C. Elphic, D.~Goldstein, D.~Summy,
  G.~D. Bart, E.~Asphaug, D.~Korycansky, D.~Landis, L.~Sollitt, Detection of
  water in the lcross ejecta plume, {\it Science\/}  463--468 (2010).

\bibitem{mitrofanov2010hydrogen}
I.~G. Mitrofanov, A.~B. Sanin, W.~Boynton, G.~Chin, J.~Garvin, D.~Golovin,
  L.~Evans, K.~Harshman, A.~Kozyrev, M.~Litvak, others, Hydrogen mapping of the
  lunar south pole using the lro neutron detector experiment lend, {\it
  Science\/}  483--486 (2010).

\bibitem{li2018direct}
S.~Li, P.~G. Lucey, R.~E. Milliken, P.~O. Hayne, E.~Fisher, J.-P. Williams,
  D.~M. Hurley, R.~C. Elphic, Direct evidence of surface exposed water ice in
  the lunar polar regions, {\it Proceedings of the National Academy of
  Sciences\/}  8907--8912 (2018).

\bibitem{scoville2022artemis}
Z.~C. Scoville, Artemis iii eva mission capability for de gerlache-shackleton
  ridge, {\it Lunar and Planetary Science Conference\/} (2022).

\bibitem{flahaut2020regions}
J.~Flahaut, J.~Carpenter, J.-P. Williams, M.~Anand, I.~Crawford, W.~van
  Westrenen, E.~F{\"u}ri, L.~Xiao, S.~Zhao, Regions of interest (roi) for
  future exploration missions to the lunar south pole, {\it Planetary and Space
  Science\/} p. 104750 (2020).

\bibitem{spudis2008geology}
P.~D. Spudis, B.~Bussey, J.~Plescia, J.-L. Josset, S.~Beauvivre, Geology of
  shackleton crater and the south pole of the moon, {\it Geophysical Research
  Letters\/} {\bf 35} (2008).

\bibitem{carrier1991physical}
W.~D. Carrier~III, G.~R. Olhoeft, W.~Mendell, Physical properties of the lunar
  surface, {\it Lunar sourcebook\/}  475--594 (1991).

\bibitem{costes1972mobility}
N.~C. Costes, J.~E. Farmer, E.~B. George, {\it Mobility Performance of the
  Lunar Roving Vehicle: Terrestrial Studies, Apollo 15 Results\/}, vol. 401
  (NASA, 1972).

\bibitem{florenskii1978floor}
C.~Florenskii, A.~Basilevskii, N.~Bobina, G.~Burba, N.~Grebennik, R.~Kuzmin,
  B.~Polosukhin, V.~Popovich, A.~Pronin, L.~Ronca, The floor of crater le
  monier-a study of lunokhod 2 data, {\it Lunar and Planetary Science
  Conference Proceedings\/},  1449--1458 (1978).

\bibitem{ding20222}
L.~Ding, R.~Zhou, Y.~Yuan, H.~Yang, J.~Li, T.~Yu, C.~Liu, J.~Wang, S.~Li,
  H.~Gao, others, A 2-year locomotive exploration and scientific investigation
  of the lunar farside by the yutu-2 rover, {\it Science Robotics\/} {\bf 7}
  (2022).

\bibitem{jones1997really}
M.~Jones, What really happened on mars rover pathfinder, {\it The Risks
  Digest\/}  1--2 (1997).

\bibitem{lindemann2006mars}
R.~A. Lindemann, D.~B. Bickler, B.~D. Harrington, G.~M. Ortiz, C.~J. Voothees,
  Mars exploration rover mobility development, {\it IEEE Robotics \& Automation
  Magazine\/}  19--26 (2006).

\bibitem{grotzinger2012mars}
J.~P. Grotzinger, J.~Crisp, A.~R. Vasavada, R.~C. Anderson, C.~J. Baker,
  R.~Barry, D.~F. Blake, P.~Conrad, K.~S. Edgett, B.~Ferdowski, others, Mars
  science laboratory mission and science investigation, {\it Space Science
  Reviews\/}  5--56 (2012).

\bibitem{farley2020mars}
K.~A. Farley, K.~H. Williford, K.~M. Stack, R.~Bhartia, A.~Chen, M.~de~la
  Torre, K.~Hand, Y.~Goreva, C.~D. Herd, R.~Hueso, others, Mars 2020 mission
  overview, {\it Space Science Reviews\/}  1--41 (2020).

\bibitem{webster2009nasa}
G.~Webster, V.~McGregor, {NASA's Mars Rover has Uncertain Future as Sixth
  Anniversary Nears},
  \url{https://mars.nasa.gov/mer/newsroom/pressreleases/20091231a.html} (2009).
  {Online; accessed 27-Jan-2023}.

\bibitem{david2005opportunity}
L.~David, Opportunity mars rover stuck in sand,
  \url{https://www.space.com/1019-opportunity-mars-rover-stuck-sand.html}
  (2005). Online; accessed 27-Jan-2023.

\bibitem{potts2015}
N.~Potts, A.~Gullikson, N.~Curran, J.~Dhaliwal, M.~Leader, R.~Rege, K.~Klaus,
  D.~Kring, {Robotic traverse and sample return strategies for a lunar farside
  mission to the Schrödinger basin}, {\it Advances in Space Research\/}
  1241--1254 (2015).

\bibitem{steenstra2016}
E.~S. Steenstra, D.~J. Martin, F.~E. McDonald, S.~Paisarnsombat, C.~Venturino,
  S.~O’Hara, A.~Calzada-Diaz, S.~Bottoms, M.~K. Leader, K.~K. Klaus, W.~{van
  Westrenen}, D.~H. Needham, D.~A. Kring, Analyses of robotic traverses and
  sample sites in the schrödinger basin for the heracles human-assisted sample
  return mission concept, {\it Advances in Space Research\/}  1050--1065
  (2016).

\bibitem{seeni2010}
A.~Seeni, B.~Schäfer, G.~Hirzinger, {Robot mobility systems for planetary
  surface exploration: state-of-the-art and future outlook: a literature
  survey}, {\it Aerospace Technologies Advancements, Chapter 10\/} p. 189–208
  (2010).

\bibitem{lee2020learning}
J.~Lee, J.~Hwangbo, L.~Wellhausen, V.~Koltun, M.~Hutter, Learning quadrupedal
  locomotion over challenging terrain, {\it Science Robotics\/} p. eabc5986
  (2020).

\bibitem{miki2022learning}
T.~Miki, J.~Lee, J.~Hwangbo, L.~Wellhausen, V.~Koltun, M.~Hutter, Learning
  robust perceptive locomotion for quadrupedal robots in the wild, {\it Science
  Robotics\/} p. eabk2822 (2022).

\bibitem{roennau2014}
A.~Roennau, G.~Heppner, M.~Nowicki, R.~Dillmann, {LAURON V: A versatile
  six-legged walking robot with advanced maneuverability}, {\it IEEE/ASME
  International Conference on Advanced Intelligent Mechatronics\/},  82--87
  (2014).

\bibitem{dirk2007bio}
S.~Dirk, K.~Frank.
\newblock The bio-inspired scorpion robot: design, control \& lessons learned.
\newblock {\it Climbing and Walking Robots: Towards New Applications\/}
  (InTech, 2007).

\bibitem{bartsch2012spaceclimber}
S.~Bartsch, T.~Birnschein, M.~R\"{o}mmermann, J.~Hilljegerdes, D.~K\"{u}hn,
  F.~Kirchner, Development of the six-legged walking and climbing robot
  spaceclimber, {\it Journal of Field Robotics\/}  506--532 (2012).

\bibitem{roennau2014reactive}
A.~Roennau, G.~Heppner, M.~Nowicki, J.~M. Z{\"o}llner, R.~Dillmann, Reactive
  posture behaviors for stable legged locomotion over steep inclines and large
  obstacles, {\it 2014 IEEE/RSJ International Conference on Intelligent Robots
  and Systems\/},  4888--4894 (IEEE, 2014).

\bibitem{kolvenbach2021martianslopes}
H.~Kolvenbach, P.~Arm, E.~Hampp, A.~Dietsche, V.~Bickel, B.~Sun, C.~Meyer,
  M.~Hutter, Traversing steep and granular martian analog slopes with a dynamic
  quadrupedal robot, {\it Field Robotics\/} (2022).

\bibitem{kolvenbach2018isairas}
H.~Kolvenbach, D.~Bellicoso, F.~Jenelten, L.~Wellhausen, M.~Hutter, {Efficient
  Gait Selection for Quadrupedal Robots on the Moon and Mars}, {\it
  International Symposium on Artificial Intelligence, Robotics and Automation
  in Space (I-SAIRAS)\/}  (2018).

\bibitem{kolvenbach2019iros}
H.~{Kolvenbach}, E.~{Hampp}, P.~{Barton}, R.~{Zenkl}, M.~{Hutter}, Towards
  jumping locomotion for quadruped robots on the moon, {\it IEEE/RSJ
  International Conference on Intelligent Robots and Systems (IROS)\/},
  5459--5466 (2019).

\bibitem{rudin2021catlike}
N.~Rudin, H.~Kolvenbach, V.~Tsounis, M.~Hutter, Cat-like jumping and landing of
  legged robots in low-gravity using deep reinforcement learning, {\it
  Transactions on Robotics\/} (IEEE, 2021).

\bibitem{arm2022results}
P.~Arm, G.~Waibel, G.~Ligeza, V.~Bickel, M.~Tranzatto, S.~Zimmermann,
  T.~Homberger, L.~Horvath, H.~Umbers, F.~Kehl, H.~Kolvenbach, M.~Hutter,
  Results and lessons learned from the first field trial of the esa-esric space
  resources challenge of team glimpse, {\it 16th Symposium on Advanced Space
  Technologies in Robotics and Automation (ASTRA 2022)\/} (2022).

\bibitem{tranzatto2022cerberus}
M.~Tranzatto, T.~Miki, M.~Dharmadhikari, L.~Bernreiter, M.~Kulkarni,
  F.~Mascarich, O.~Andersson, S.~Khattak, M.~Hutter, R.~Siegwart, K.~Alexis,
  Cerberus in the darpa subterranean challenge, {\it Science Robotics\/} p.
  eabp9742 (2022).

\bibitem{hudson2021heterogeneous}
N.~Hudson, F.~Talbot, M.~Cox, J.~Williams, T.~Hines, A.~Pitt, B.~Wood,
  D.~Frousheger, K.~L. Surdo, T.~Molnar, others, Heterogeneous ground and air
  platforms, homogeneous sensing: Team csiro data61's approach to the darpa
  subterranean challenge, {\it arXiv preprint arXiv:2104.09053\/}  (2021).

\bibitem{agha2021nebula}
A.~Agha, K.~Otsu, B.~Morrell, D.~D. Fan, R.~Thakker, A.~Santamaria-Navarro,
  S.-K. Kim, A.~Bouman, X.~Lei, J.~Edlund, others, Nebula: Quest for robotic
  autonomy in challenging environments; team costar at the darpa subterranean
  challenge, {\it arXiv preprint arXiv:2103.11470\/}  (2021).

\bibitem{schuster2020arches}
M.~J. Schuster, M.~G. M{\"u}ller, S.~G. Brunner, H.~Lehner, P.~Lehner,
  R.~Sakagami, A.~D{\"o}mel, L.~Meyer, B.~Vodermayer, R.~Giubilato, others, The
  arches space-analogue demonstration mission: Towards heterogeneous teams of
  autonomous robots for collaborative scientific sampling in planetary
  exploration, {\it IEEE Robotics and Automation Letters\/}  5315--5322 (2020).

\bibitem{cordes2011lunares}
F.~Cordes, I.~Ahrns, S.~Bartsch, T.~Birnschein, A.~Dettmann, S.~Estable,
  S.~Haase, J.~Hilljegerdes, D.~Koebel, S.~Planthaber, others, Lunares: Lunar
  crater exploration with heterogeneous multi robot systems, {\it Intelligent
  Service Robotics\/}  61--89 (2011).

\bibitem{sonsalla2017field}
R.~Sonsalla, F.~Cordes, L.~Christensen, T.~M. Roehr, T.~Stark, S.~Planthaber,
  M.~Maurus, M.~Mallwitz, E.~A. Kirchner, Field testing of a cooperative
  multi-robot sample return mission in mars analogue environment, {\it 14th
  Symposium on advanced space technologies in robotics and automation
  (ASTRA)\/} (2017).

\bibitem{nasa2022braille}
B.~Dynamics, {Search for Life: NASA JPL Explores Martian-Like Caves},
  \url{https://www.youtube.com/watch?v=qTW-dbZr4U8} (2022). Online; accessed
  23-January-2023.

\bibitem{tzanetos2022ingenuity}
T.~Tzanetos, M.~Aung, J.~Balaram, H.~F. Grip, J.~T. Karras, T.~K. Canham,
  G.~Kubiak, J.~Anderson, G.~Merewether, M.~Starch, others, Ingenuity mars
  helicopter: From technology demonstration to extraterrestrial scout, {\it
  2022 IEEE Aerospace Conference (AERO)\/},  01--19 (IEEE, 2022).

\bibitem{hutter2017anymal}
M.~Hutter, C.~Gehring, A.~Lauber, F.~Gunther, C.~D. Bellicoso, V.~Tsounis,
  P.~Fankhauser, R.~Diethelm, S.~Bachmann, M.~Bloesch, H.~Kolvenbach,
  M.~Bjelonic, L.~Isler, K.~Meyer, {ANYmal - toward legged robots for harsh
  environments}, {\it Advanced Robotics\/}  918--931 (2017).

\bibitem{oravec2021geotechnical}
H.~A. Oravec, V.~M. Asnani, C.~M. Creage, S.~J. Moreland.
\newblock Geotechnical review of existing mars soil simulants for surface
  mobility.
\newblock {\it Earth and Space 2021\/} (2021),  157--170.

\bibitem{kolvenbach2018scalability}
H.~Kolvenbach, M.~Breitenstein, C.~Gehring, M.~Hutter, Scalability analysis of
  legged robots for space exploration, {\it 68th International Astronautical
  Congress (IAC 2017)\/},  10399--10413 (Curran, 2018).

\bibitem{valsecchi2022preliminary}
G.~Valsecchi, D.~Liconti, F.~Tischhauser, H.~Kolvenbach, M.~Hutter, Preliminary
  design of actuators for walking robot on the moon, {\it 16th Symposium on
  Advanced Space Technologies in Robotics and Automation (ASTRA 2022)\/}
  (2022).

\bibitem{mira}
Metrohm, Metrohm mira xtr ds,
  \url{https://www.metrohm.com/en/products/raman-spectroscopy/mira-ds-mira-xtr-ds.html}
  (2023). Online; accessed 23-January-2023.

\bibitem{peng2020learning}
X.~B. Peng, E.~Coumans, T.~Zhang, T.-W. Lee, J.~Tan, S.~Levine, Learning agile
  robotic locomotion skills by imitating animals, {\it arXiv preprint
  arXiv:2004.00784\/}  (2020).

\bibitem{tranzatto2022team}
M.~Tranzatto, M.~Dharmadhikari, L.~Bernreiter, M.~Camurri, S.~Khattak,
  F.~Mascarich, P.~Pfreundschuh, D.~Wisth, S.~Zimmermann, M.~Kulkarni, others,
  Team cerberus wins the darpa subterranean challenge: Technical overview and
  lessons learned, {\it arXiv preprint arXiv:2207.04914\/}  (2022).

\bibitem{compSLAM}
S.~Khattak, H.~Nguyen, F.~Mascarich, T.~Dang, K.~Alexis, Complementary
  multi--modal sensor fusion for resilient robot pose estimation in
  subterranean environments, {\it 2020 International Conference on Unmanned
  Aircraft Systems (ICUAS)\/},  1024--1029 (IEEE, 2020).

\bibitem{p2planeICP}
K.-L. Low, Linear least-squares optimization for point-to-plane icp surface
  registration, {\it Chapel Hill, University of North Carolina\/}  1--3 (2004).

\bibitem{kazhdan2006poisson}
M.~Kazhdan, M.~Bolitho, H.~Hoppe, Poisson surface reconstruction, {\it
  Proceedings of the fourth Eurographics symposium on Geometry processing\/},
  {\bf 7} (2006).

\bibitem{open3d}
Q.-Y. Zhou, J.~Park, V.~Koltun, Open3d: A modern library for 3d data
  processing, {\it arXiv preprint arXiv:1801.09847\/}  (2018).

\bibitem{edl}
C.~Boucheny, Interactive scientific visualization of large datasets: towards a
  perceptive-based approach, Ph.D. thesis, Universit{\'e} Joseph Fourier
  Grenoble, France (2009).

\bibitem{draco}
Google, Draco 3d graphics compression, \url{https://google.github.io/draco/}
  (2017). Online; accessed 27-January-2023.

\bibitem{miki2022elevation}
T.~Miki, L.~Wellhausen, R.~Grandia, F.~Jenelten, T.~Homberger, M.~Hutter,
  Elevation mapping for locomotion and navigation using gpu, {\it 2022 IEEE/RSJ
  International Conference on Intelligent Robots and Systems (IROS)\/},
  2273--2280 (IEEE, 2022).

\bibitem{he2017mask}
K.~He, G.~Gkioxari, P.~Doll{\'a}r, R.~Girshick, Mask r-cnn, {\it Proceedings of
  the IEEE international conference on computer vision\/},  2961--2969 (2017).

\bibitem{faconti2019mood2be}
D.~Faconti, Mood2be: Models and tools to design robotic behaviors, {\it Eurecat
  Centre Tecnologic, Barcelona, Spain, Tech. Rep\/} {\bf 4} (2019).

\bibitem{wellhausen2021rough}
L.~Wellhausen, M.~Hutter, Rough terrain navigation for legged robots using
  reachability planning and template learning, {\it 2021 IEEE/RSJ International
  Conference on Intelligent Robots and Systems (IROS)\/},  6914--6921 (IEEE,
  2021).

\bibitem{rajant}
Rajant, \url{https://rajant.com/products/breadcrumb-wireless-nodes/dx-series/}
  (2023). Online; accessed 09-January-2023.

\bibitem{quigley2009ros}
M.~Quigley, K.~Conley, B.~Gerkey, J.~Faust, T.~Foote, J.~Leibs, R.~Wheeler,
  A.~Y. Ng, Ros: an open-source robot operating system, {\it ICRA workshop on
  open source software\/}, p.~5 (Kobe, Japan, 2009).

\bibitem{stuckler2016nimbro}
J.~St{\"u}ckler, M.~Schwarz, M.~Schadler, A.~Topalidou-Kyniazopoulou,
  S.~Behnke, Nimbro explorer: Semiautonomous exploration and mobile
  manipulation in rough terrain, {\it Journal of Field Robotics\/}  411--430
  (2016).

\bibitem{schwarz2017nimbro}
M.~Schwarz, T.~Rodehutskors, D.~Droeschel, M.~Beul, M.~Schreiber, N.~Araslanov,
  I.~Ivanov, C.~Lenz, J.~Razlaw, S.~Sch{\"u}ller, D.~Schwarz,
  A.~Toplaidou-Kyniazopoulou, S.~Behnke, Nimbro rescue: Solving
  disaster-response tasks with the mobile manipulation robot momaro, {\it
  Journal of Field Robotics\/}  400--425 (2017).

\bibitem{kaupisch2015dlr}
T.~Kaupisch, D.~Noelke, A.~Arghir, Dlr spacebot cup—germany’s space
  robotics competition, {\it Symposium on Advanced Space Technologies in
  Robotics and Automation (ASTRA)\/} (2015).

\bibitem{lee2023learning}
S.~Lee, S.~Jeon, J.~Hwangbo, Learning legged mobile manipulation using
  reinforcement learning, {\it Robot Intelligence Technology and Applications
  7: Results from the 10th International Conference on Robot Intelligence
  Technology and Applications\/},  310--317 (Springer, 2023).

\bibitem{deepwbc}
Z.~Fu, X.~Cheng, D.~Pathak, Deep whole-body control: Learning a unified policy
  for manipulation and locomotion, {\it Conference on Robot Learning
  ({CoRL})\/} (2022).

\bibitem{ma2022combining}
Y.~Ma, F.~Farshidian, T.~Miki, J.~Lee, M.~Hutter, Combining learning-based
  locomotion policy with model-based manipulation for legged mobile
  manipulators, {\it IEEE Robotics and Automation Letters\/}  2377--2384
  (2022).

\bibitem{ocs2}
F.~Farshidian, R.~Grandia, M.~Spieler, J.~Carius, J.-P. Sleiman, Ocs2,
  \url{https://leggedrobotics.github.io/ocs2/} (2022). Online; accessed
  27-September-2022.

\bibitem{chiu2022collision}
J.-R. Chiu, J.-P. Sleiman, M.~Mittal, F.~Farshidian, M.~Hutter, A
  collision-free mpc for whole-body dynamic locomotion and manipulation, {\it
  2022 International Conference on Robotics and Automation (ICRA)\/},
  4686--4693 (IEEE, 2022).

\bibitem{DBLP:journals/corr/abs-2103-00946}
J.~Sleiman, F.~Farshidian, M.~V. Minniti, M.~Hutter, A unified {MPC} framework
  for whole-body dynamic locomotion and manipulation, {\it CoRR\/}  (2021).

\bibitem{DBLP:journals/corr/abs-2103-15980}
J.~L. Blanco{-}Claraco, A tutorial on se(3) transformation parameterizations
  and on-manifold optimization, {\it CoRR\/}  (2021).

\bibitem{miraControl}
F.~Abi-Farraj, N.~Pedemonte, P.~Robuffo~Giordano, A visual-based shared control
  architecture for remote telemanipulation, {\it 2016 IEEE/RSJ International
  Conference on Intelligent Robots and Systems (IROS)\/},  4266--4273 (2016).

\bibitem{behaviortreeCpp}
D.~Faconti, Behaviortree.cpp, \url{https://www.behaviortree.dev/} (2022).
  Online; accessed 05-October-2022.

\bibitem{Sojourner}
JPL, A description of the rover sojourner,
  \url{https://mars.nasa.gov/MPF/rover/descrip.html}. Online; accessed
  27-January-2023.

\bibitem{MER}
NASA, Mars exploration rovers overview,
  \url{https://mars.nasa.gov/mer/mission/rover/wheels-and-legs/}. Online;
  accessed 27-January-2023.

\bibitem{ding2022surface}
L.~Ding, R.~Zhou, T.~Yu, H.~Gao, H.~Yang, J.~Li, Y.~Yuan, C.~Liu, J.~Wang,
  Y.~Zhao, others, Surface characteristics of the zhurong mars rover traverse
  at utopia planitia, {\it Nature Geoscience\/}  171--176 (2022).

\bibitem{heverly2010development}
M.~Heverly, J.~Matthews, M.~Frost, C.~Quin, Development of the tri-athlete
  lunar vehicle prototype, {\it Proceedings of the 40th Aerospace Mechanisms
  Symposium\/} (2010).

\bibitem{CREX}
DFKI, Crex: Crater explorer,
  \url{https://robotik.dfki-bremen.de/en/research/robot-systems/crex/}. Online;
  accessed 27-January-2023.

\bibitem{bartsch2012development}
S.~Bartsch, T.~Birnschein, M.~R{\"o}mmermann, J.~Hilljegerdes, D.~K{\"u}hn,
  F.~Kirchner, Development of the six-legged walking and climbing robot
  spaceclimber, {\it Journal of Field Robotics\/}  506--532 (2012).

\bibitem{arm2019spacebok}
P.~Arm, R.~Zenkl, P.~Barton, L.~Beglinger, A.~Dietsche, L.~Ferrazzini,
  E.~Hampp, J.~Hinder, C.~Huber, D.~Schaufelberger, F.~Schmitt, B.~Sun,
  B.~Stolz, H.~Kolvenbach, M.~Hutter, {SpaceBok: A Dynamic Legged Robot for
  Space Exploration}, {\it IEEE International Conference on Robotics and
  Automation (ICRA)\/} (2019).

\bibitem{playter2006bigdog}
R.~Playter, M.~Buehler, M.~Raibert, Bigdog, {\it Unmanned Systems Technology
  VIII\/},  896--901 (SPIE, 2006).

\bibitem{Spot}
B.~Dynamics, Spot specifications,
  \url{https://support.bostondynamics.com/s/article/Robot-specifications}.
  Online; accessed 27-January-2023.

\bibitem{ANYmal}
ANYbotics, Anymal specifications,
  \url{https://www.anybotics.com/anymal-autonomous-legged-robot/}. Online;
  accessed 27-January-2023.

\bibitem{katz2019mini}
B.~Katz, J.~Di~Carlo, S.~Kim, Mini cheetah: A platform for pushing the limits
  of dynamic quadruped control, {\it 2019 international conference on robotics
  and automation (ICRA)\/},  6295--6301 (IEEE, 2019).

\end{thebibliography}

\clearpage
\newpage

\setcounter{table}{0}
\makeatletter 
\renewcommand{\thetable}{S\@arabic\c@table}
\makeatother

\setcounter{figure}{0}
\makeatletter 
\renewcommand{\thefigure}{S\@arabic\c@figure}
\makeatother

\setcounter{algorithm}{0}
\makeatletter 
\renewcommand{\thealgorithm}{S\@arabic\c@algorithm}
\makeatother

\section*{Supplementary Results}

\subsection*{Power Consumption Experiments}
To compare the energy efficiency of our controller and the baseline controller~\cite{miki2022learning}, we conducted a mock-mission experiment in a laboratory setting with the Scientist robot (Movie~S2). We pre-defined two waypoints for the robot, \SI{3.8}{\meter} apart, and let the robot walk back and forth between the waypoints with a static arm. The robot stopped for \SI{8}{\second} at every waypoint, simulating a short measurement. We powered the robot via an EPS PSB 9000 3U bidirectional power supply that allows measuring the power consumption of the whole robot directly via a serial interface.

During standby, meaning with all computers and actuators running in standby mode, the robot consumes \SI{175}{\watt}. We subtracted this value from all subsequent measurements to isolate the additional locomotion power consumption. This includes the actuator power consumption, but also increased power consumption and losses in other subsystems, such as the computers, fans, and power electronics. During the mission with our controller, the robot had an average locomotion power consumption of \SI{403}{W}, whereas the baseline controller resulted in a locomotion power consumption of \SI{475}{\watt}, which corresponds to a \SI{15}{\percent} improvement with our controller.

We additionally isolated the walking sequences from our mission and computed the average locomotion power consumption per walking sequence. With 40 waypoints, this resulted in 40 samples per controller. The means of the datasets are \SI{808}{\watt} and \SI{749}{\watt}, using the our controller and the baseline controller, respectively. Even though the improvement is smaller without the standing sequences, the gain is still substantial (Fig.~\ref{fig:PowerConsumption}).

\section*{Supplementary Methods}
\subsection*{Locomotion Policy Details}
\label{sec:locomotion_details}
In this section, we present the implementation details of the training for the locomotion controllers deployed on the robots. Our controllers are based on our previous work~\cite{miki2022learning}. However, heavy and potentially moving equipment, such as robotic arms and scientific payloads, can impede the robustness and efficiency of the locomotion controller. Other researchers have designed reinforcement learning based locomotion policies for legged mobile manipulators~\cite{lee2023learning, deepwbc}, but these approaches were only validated in simulation~\cite{lee2023learning} or on flat terrain~\cite{deepwbc}.

To achieve robust locomotion with an additional robotic arm, we added the arm to the robot in simulation, providing realistic disturbances and explicitly included arm joint position and velocity measurements in the observation, as opposed to the random wrench used during training in our previous work~\cite{ma2022combining}. Compared to the previous work, where the wrench is predicted based on the MPC plan of the arm controller during deployment, we only rely on the measured arm joint positions and velocities as observations, which improves the robot's robustness. The observations of our baseline~\cite{miki2022learning} and the modified controller are shown in Table~S6.

Additionally, our baseline~\cite{miki2022learning} did not prioritize energy efficiency, resulting in inefficient locomotion with additional payloads, as shown in the Supplementary Results. To address this, we incorporated payload mass randomization and added the mass information as privileged information to the teacher policy. This enables the policy to compensate for the additional weight more effectively. Without this change, the previous controller walked at a low body height, which requires larger actuator torques. Moreover, we increased the torque penalty reward to further improve locomotion efficiency.

The arm joint in the simulation is controlled by a PD controller to follow a randomly sampled target joint position independently of the locomotion controller (Fig.~\ref{fig:LocomotionTraining}) while the policy controls only the leg joints.

To make the training more stable with the additional randomization, we used the following curriculum on top of the previous work~\cite{miki2022learning}. 
During training, the curriculum factor $c_k$ was updated exponentially every training episode $c_{k+1} = c_k^d$, with convergence rate $d = 0.98$. We used $c_0 = 0.07$ as the initial value.
We increased the mass of the arm and the randomized payload mass based on the curriculum factor, 
\begin{eqnarray}
m_{arm}^{sim} &=& m_{arm} \cdot c_k \\
m_{payload} &=& ~\mathcal{U}(-2.0, 7.0) \cdot c_k,
\end{eqnarray}
where $m_{arm}$ is the original mass of the arm, $m_{arm}^{sim}$ is the mass applied in the physics simulation, and $\mathcal{U}$ represents the uniform distribution.
Additionally, we also varied the range of the initial and target arm positions based on the curriculum, 
\begin{eqnarray}
q_{arm}^{initial} = q_{arm}^{nominal} + 2.2 \cdot \mathcal{N}(0, c_k) \\
q_{arm}^{target} = q_{arm}^{nominal} + 2.2 \cdot \mathcal{N}(0, c_k),
\end{eqnarray}
where $q_{arm}$ represents the joint position of the arm and $\mathcal{N}$ represents the normal distribution.
These values are updated every training episode.

\subsection*{Arm Control Implementation Details}
\label{sec:ArmControl}

For the arm control, we model the robot as a floating-base 6 DoF manipulator using the OCS2 library \cite{ocs2}. We encode all cost terms for the arm control in a single cost function according to \cite{chiu2022collision} as follows: 
\begin{equation}
\begin{split}
    L(\mathbf{x},\mathbf{u},t) = &\alpha_1  \underbrace{|| \mathbf{r}_{IE} - \mathbf{r}_{IE}^{ref}||^2_{\mathbf{Q}_{ee}}}_\text{Tool pose tracking cost} + \alpha_2 \underbrace{|| \mathbf{x}_{r} - \mathbf{x}_{r}^{ref}||^2_{\mathbf{Q}_{r}}}_{\substack{\text{Nominal state} \\ 
    \text{deviation cost}}} \\
    &+ \underbrace{|| \mathbf{u} - \mathbf{u}^{ref}||^2_{\mathbf{R}}}_\text{Input cost} + \underbrace{L_c(\mathbf{x}_r, t)}_\text{Self-collision cost},
\end{split}
\end{equation}
where $\mathbf{r}_{IE} \in {\rm I\!R}^3$ denotes the tool pose in the inertial control frame, $\mathbf{x}_r = (\mathbf{q}_b, \mathbf{q}_j) \in {\rm I\!R}^{13}$ denotes the robot state that is composed of the base pose and the arm joint positions, and $\mathbf{u} \in {\rm I\!R}^6$ denotes the input to the system composed of the joint velocities. The weight matrices $\mathbf{Q}_{ee}$ and $\mathbf{Q}_r$ are positive semi-definite, and $\mathbf{R}$ is positive definite. The parameters $\alpha_1, \alpha_2 \in \{0,1\}$ are used to determine whether a cost term is active or not according to the current task. In this work, we always use $\alpha_2=1$. During walking, we set large weights in the matrix $Q_r$ to keep the arm stable in an efficient nominal joint configuration with a low center of gravity and disable the end effector tracking term by setting $\alpha_1=0$. During measurement tasks, we set small weights in the matrix $Q_r$ and we enable the end-effector tracking task by setting $\alpha_2=1$. Finally, $L_c(\mathbf{x}_r, t)$ denotes the penalties corresponding to the self-collision constraints. Compared to \cite{DBLP:journals/corr/abs-2103-00946}, we use a hybrid control approach where we employ separate control approaches for manipulation and locomotion. We run the MPC controller for the arm and our reinforcement-learning-based locomotion policy for the base instead of running a whole-body MPC controller.

The arm control has a pose command interface and a twist command interface. The pose command interface directly updates $\mathbf{r}_{IE}^{ref}$. The twist command interface receives a desired translational $\mathbf{v}^{ref}$ and angular $\boldsymbol{\omega}^{ref}$ task space velocity. The desired pose $\mathbf{r}_{IE}^{ref} = (\mathbf{p}_{IE}^{ref}, \mathbf{q}_{IE}^{ref})$ is then updated with the time step $dt$ of the controller using

\begin{eqnarray}
\mathbf{p}_{IE}^{ref} = \mathbf{p}_{IE} + \mathbf{v}^{ref} dt \nonumber \\
\mathbf{q}_{IE}^{ref} = \mathbf{q}_{IE}  \boxplus  (\boldsymbol{\omega}_{IE}^{ref} dt)
\end{eqnarray}

Where the $\boxplus$ operator is the additive operator for Euclidean spaces \cite{DBLP:journals/corr/abs-2103-15980}. It can be expressed as multiplying $\mathbf{q}_{IE}$ with the exponential map of the vector $\boldsymbol{\omega}_{IE}^{ref}$ written as $e^{\boldsymbol{\omega}_{IE}^{ref} dt}$.

To take a measurement with the MIRA, the sensor needs to aim at the target. The optimal distance for the stand-off attachment $d_{desired}$ is \SI{0.7}{\metre}. We use an iterative algorithm to minimize the alignment error between the sensor and the target. Abi Farraj et al. \cite{miraControl} implemented a control law to keep the gripper aligned with the target next to other objectives. According to this method we calculate the instantaneous alignment error between the sensor and the target with the desired distance $d_{desired}$. 

\begin{eqnarray}
\mathbf{r}_{error} = (\mathbf{r}_{alignment} / ||\mathbf{r}_{tool-target}||)(\mathbf{r}_{tool-target}\mathbf{r}_{tool-target}^T - \mathbf{\textit{I}})^{-1} \nonumber \\
+ ( ||\mathbf{r}_{tool-target}|| - d_{desired})\mathbf{r}_{tool-target}
\end{eqnarray}

where $\mathbf{r}_{alignment}$ denotes the unit vector along the Raman laser. All vectors $\mathbf{r}_{alignment}$, $\mathbf{r}_{error}$ and $
\mathbf{r}_{tool-target}$ are position vectors in ${\rm I\!R}^3$ calculated in the tool frame (Fig.~\ref{fig:MiraDeployment}). Secondly, a PID control law transforms the error vector into a desired instantaneous velocity for the end effector. Finally, the end effector velocity is transformed into the control frame of the robot and sent to the robotic arm controller using the twist interface. This loop is executed until the magnitude of the alignment error $\mathbf{r}_{error}$ is  lower than \SI{0.01}{\meter}.

The first stage of the MICRO deployment uses the controller's pose interface to directly send the desired MICRO pose to the arm controller. The second stage uses a PID control loop to calculate the desired twist:

\begin{equation}
  \mathbf{v}^{ref} = K_p \mathbf{e} + K_d \mathbf{v} + K_i \int \mathbf{e},
\end{equation}

where $\mathbf{v}^{ref}$ denotes the reference velocity in the tool frame, $\mathbf{v}$ denotes the current tool velocity in the tool frame, $K_p$, $K_d$ and $K_i$ denote the gains of the PID controller, and $\mathbf{e} = (-r_x^{target}, -r_y^{target}, e_z)$ denotes the error vector. $r_x^{target}$ and $r_y^{target}$ are defined by the operator via the marker interface (Sec.~Autonomous In-Situ Measurement Acquisition) and $e_z = d_{Tof} - d^{ref}$, denotes the difference between distance measured MICRO's time of flight sensor (Fig.~S1) and the desired distance between the time of flight sensor and the target surface. We select $d^{ref}$ to compress the soft foam at MICRO's front to block ambient light without damaging MICRO. We found the best results with $d^{ref} =  \SI{0.005}{\meter}$.  Finally, the reference velocity is transformed from the tool frame to the inertial control frame and sent to the robotic arm controller using the twist interface.

\subsection*{Behavior Trees for Robust Task-Level Autonomy}
\label{sec:behavior_tree}
We use behavior trees to create complex behaviors that are modular, reactive, and reusable. Behavior trees are directed, acyclic graphs. Each leaf has only one parent, which makes transitions between actions easy in contrast to the tight coupling of states in finite state machines (FSMs). Tight coupling makes it difficult to add, remove, or reuse a state because it requires changing the conditions of all other states that contain transitions to the new or old state. Therefore, in our experience, behavior trees are easier to adapt when implementing a large and complex system.

The modularity of behavior trees allows us to reuse many subtrees across our team of robots, as shown in Fig.~\ref{fig:AllRobotBT}. All our robots share the same subtree to perform navigation tasks. In addition, we have subtrees to autonomously execute remote and in-situ measurement tasks (called "inspection" and "measurement" in our behavior tree framework, ~Fig.~S13). The individual subtrees have a large overlap across the robots and are minimally adapted to the specific payload setup of each robot. We illustrate this with the measurement subtree: Although they are largely the same for the Hybrid and Scientist, the Hybrid lacks the MICRO and the ability to point the instruments with the robotic arm. Therefore, Hybrid's measurement subtree does not contain the "MicroPointSubtree", "MicroMeasurementSubtree" and the "MiraPointingSubtree".

The Scientist's behavior tree, which is the largest one, has 76 leaf nodes. This includes check, action, and monitoring nodes but does not include control or decorator nodes, which are used to control the data flow through the tree structure.

During the mission, the operator can monitor the current state of the behavior tree and specific information about requested actions. The operator can cancel a running action or send another action. A new action request cancels the running action and starts the requested action. For certain actions, the operator can specify non-default parameters. For example, the operator can request a CTX-FW point inspection with a specific filter instead of all filters.

We use Behaviortree.CPP \cite{behaviortreeCpp} to implement our behavior trees. It provides a ROS interface suitable for our application and basic implementations for action, condition, and decorator nodes that can be customized to generate specialized behaviors. Furthermore, Behaviortree.CPP includes a graphical user interface called Groot to facilitate the development of complex tree structures. There is no distinction between data flow and control flow. Data is passed through the nodes as the action is executed.

\renewcommand{\thetable}{S\arabic{table}}
\setcounter{table}{0}

\begin{table*}
\caption{\textbf{Redundancy Visualization for Scientific Instrument Deployment} This table is a simplified visualization: There is additional redundancy between instruments. For example, MICRO and CTX-FW both provide multispectral imaging (but on different scales). \checkmark: Task can be fully executed. $\circ$: Task can be executed with limited quality. \ding{55}: Task cannot be executed.}
    \centering
    \begin{tabular}{c|c|c||c|c|c|c|c}
        \hline
        \multicolumn{3}{c||}{Inactive Robots} & \multicolumn{5}{c}{Scientific Instrument Deployment Task} \\
        \hline
        Scout          &Hybrid         &Scientist      &CTX                &FW             &TH             &MIRA           &MICRO\\
        \hline
                       &               &               &\checkmark        &\checkmark    &\checkmark    &\checkmark     &\checkmark\\
        \hline
        \ding{55}    &               &               &\checkmark        &\ding{55}    &\checkmark    &\checkmark     &\checkmark\\
        \hline
                       &\ding{55}    &               &\checkmark        &\checkmark    &\ding{55}    &\checkmark     &\checkmark\\
        \hline
                       &                &\ding{55}   &\checkmark        &\checkmark    &\checkmark    &$\circ$    &\ding{55}\\
        \hline
        \ding{55}    &\ding{55}     &              &\ding{55}        &\ding{55}    &\ding{55}    &\checkmark     &\checkmark\\
        \hline
        \ding{55}    &                &\ding{55}   &\checkmark        &\ding{55}    &\checkmark    &$\circ$    &\ding{55}\\
        \hline
                       &\ding{55}     &\ding{55}   &\checkmark        &\checkmark    &\ding{55}    &\ding{55}     &\ding{55}\\
        \hline
        \ding{55}    &\ding{55}     &\ding{55}   &\ding{55}        &\ding{55}    &\ding{55}    &\ding{55}     &\ding{55}\\
        \hline
    \end{tabular}
    \label{tab:Redundancy}
\end{table*}

\begin{table*}
\caption{\textbf{Overview of important Space Resources Challenge objectives and conditions}}
\centering
\begin{tabular}{l}
\hline
\textbf{Objectives} \\
\hline
Find and map the locations of resource-enriched areas (REAs) \\       
Characterize the composition of the REAs \\
Find and map boulders and obstacles \\
Characterize the composition of the boulders \\
Create a digital elevation map of the area \\
Demonstrate operations during loss of signal \\
Demonstrate a user-friendly interface \\
Demonstrate scalability and fault tolerance \\
Demonstrate safe operations and obstacle avoidance \\
Write a report containing all results \\
\hline
\textbf{Conditions} \\
\hline
The ground is basaltic \\
The slope does not exceed \SI{25}{\degree} \\
No overview map is provided before the challenge, but the overall area dimensions are given\\
The light is at a steep incidence angle (analog to the lunar south pole) \\
The total mass of the assets must not exceed \SI{300}{\kilo \gram} \\
A single asset must not exceed \SI{100}{\kilo \gram}\\
Navigation cannot rely on Earth's magnetic field or GPS \\
One payload can be placed at the lander with a power and a network connection \\
Downlink and uplink are \SI{100}{\mega \bit \per \second} \\
The communication between the challenge area and the outside has a \SI{5}{\second} rount-trip-time\\
One loss of signal occurs at a pre-defined time \\
An additional loss of signal occurs at a random time \\
Five team members are allowed in the mission control room \\
The team has five hours to exectue the mission and write the report \\
\hline
\label{tab:SRC_Rules}
\end{tabular}
\end{table*}

\begin{sidewaystable*}
\caption{\textbf{Specification of different planetary exploration rovers and research platforms concerning maximum speed and slope climbing capabilities.} N/A: no data available.}
    \centering
    \begin{tabular}{ll|c|c}
        \hline
        \multicolumn{2}{l|}{Robots}                                 &  Maximum speed capability on flat ground    &  Slope climbing capability           \\
        \multicolumn{2}{l|}{}                                       &  (mm/s) / (body length/s)                     &  (deg)              \\
        \hline
        \multicolumn{2}{l|}{\textbf{Planetary exploration rovers}}  &                              &                                      \\
        \multicolumn{2}{l|}{Sojourner \cite{Sojourner}}             &  6.7 / 0.011                       &  N/A                          \\
        \multicolumn{2}{l|}{Spirit and Opportunity \cite{MER}}      &  50 / 0.031                         &  17                           \\
        \multicolumn{2}{l|}{Curiosity}                              &  44 / 0.015                         &  N/A                          \\
        \multicolumn{2}{l|}{Yutu-2 \cite{ding20222}}                &  56 / 0.037                         &  20                           \\
        \multicolumn{2}{l|}{Perseverance}                           &  44.7 / 0.015                       &  N/A                          \\
        \multicolumn{2}{l|}{Zhurong \cite{ding2022surface}}         &  56 / 0.026                         &  30 (on rigid ground)         \\
        \multicolumn{2}{l|}{}                                       &                                     &  20 (on soft ground)          \\
        \hline
        \multicolumn{2}{l|}{\textbf{Wheeled-leg hybrid robots}}     &                                     &                               \\
        \multicolumn{2}{l|}{Athlete \cite{heverly2010development}}  &  833 / N/A                               &  up to 10                     \\
        \hline
        \multicolumn{2}{l|}{\textbf{Legged robots}}             &                                                      &                                 \\
        CREX \cite{CREX}                                    & \multirow{5}*{Planetary Prototypes}      &  120 / 0.146                                                &  N/A   \\
        SpaceClimber \cite{bartsch2012spaceclimber}          & ~ &  175 / 2.500                                         &  35 (in artificial lunar crater)          \\
                                                            & ~ &                                                      &  25 (on ramp)                             \\
        SpaceBok \cite{arm2019spacebok,kolvenbach2021martianslopes} & ~ &  1000 / 1.667                                      &  min 25 on granular soil                       \\ 
        ANYmal (ours) \cite{miki2022learning}    & ~ &  1900 / 2.043                          &  min 25 on granular soil                          \\
        \hline
        Big Dog \cite{playter2006bigdog}      & \multirow{5}*{Terrestrial}   &  700 / 0.700 (walk over loose rock beds)       &  25 (walk up)     \\
                                                            & ~ &  1800 / 1.800 (trot)                    &  35 (walk down)                                      \\
        Spot  \cite{Spot}                     & ~ &  1600 / 1.455                          &  30                                                  \\
        ANYmal (off-the-shelf) \cite{ANYmal}                   & ~ &  1300 / 1.400                          &  30                                                  \\
        Mini Cheetah \cite{katz2019mini} & ~ &  2450 / 6.125                          &  N/A                                                 \\
        Unitree Go1 Pro                       & ~ &  4700 / 7.993                          &  N/A                                                 \\
        \hline
    \end{tabular}
    \label{tab:LocomotionCapability}
\end{sidewaystable*}

\begin{table*}
\caption{\textbf{Computing Power on customized ANYmal robots}. \\
The right column shows a non-exhaustive list of major software components executed on the respective computer.}
\centering
\begin{tabular}{l|ll}
\hline
Type                        & Main software modules \\
\hline               
Intel core i7 8850H with 2 x \SI{8}{\giga \byte} memory            & State Estimation, Locomotion Controller, \\ 
                                                                    & Hardware Interfaces, Communication \\
\hline
Intel core i7 8850H with 2 x \SI{8}{\giga \byte} memory            & Localization, Dense Mapper, Local Planner, \\ 
                                                                    &Behavior Tree, Hardware Interfaces  \\
\hline
Jetson AGX Xavier (Scout, Scientist),             & Elevation Mapping, Image Processing \\
Jetson AGX Orin (Hybrid) \\
\label{tab:ComputingPower}
\end{tabular}
\end{table*}

\begin{table*}
\caption{\textbf{List of minerals in the MIRA Database}. \\
All reference measurements were acquired from \SI{0.4}{m} distance to the sample with the Metrohm Mira XTR DS}
\centering
\begin{tabular}{l|ll}
\hline
Mineral group and name      & Number of reference spectra \\
\hline
\textbf{Oxide Group}        & \\
Ilmenite                    & 30 \\
Rutile                      & 30  \\
Titanium dioxide (TiO2)     & 30 \\
Spinel                      & 15 \\
Chromite                    & 15 \\
Magnetite                   & 15 \\
\hline
\textbf{Olivine Group}      & \\
Forsterite                  & 15 \\
\hline
\textbf{Pyroxene Group}     & \\
Augite                      & 15 \\
Aegirine                    & 16 \\
Diopside                    & 15 \\
Enstatite                   & 16 \\
Hedenbergite                & 15 \\
\hline
\textbf{Feldspar Group}     & \\
Albite                      & 16 \\
Andesine                    & 15 \\
Anorthite                   & 15 \\
Anorthoclase                & 15 \\
Bytownite                   & 16 \\
Orthoclase                  & 15 \\
Oligoclase                  & 15 \\
\hline
\textbf{Lunar Regolith Simulant} & 21 \\
\hline
\textbf{Total}              & 355
\label{tab:MiraDatabase}
\end{tabular}
\end{table*}

\begin{table*}[htbp]
\caption{\textbf{Comparison of original~\cite{miki2022learning} and modified observations.} Proprioception is used for both teacher and student training. Exteroception is given in the form of height samples. The privileged information is used only for teacher training.}
\centering

\begin{tabular}{l|ll|ll}
\hline
                 & Original \cite{miki2022learning}                    &                &  Modified    & \\ \hline
Observation type & Input                       & Dim            &  input       & Dim\\ \hline
Proprioception   & command                     & 3              & command                      &   3  \\
                 & body orientation            & 3              & body orientation             &   3  \\
                 & body velocity               & 6              & body velocity                &   6  \\
                 & joint position              & 12             & \textbf{leg + arm} joint position  & \textbf{18}\\
                 & joint velocity              & 12             & \textbf{leg + arm} joint velocity               &   \textbf{18} \\
                 & joint position history      & 36             & \textbf{leg + arm} joint position history       &   \textbf{54} \\
                 & joint velocity history      & 24             & \textbf{leg + arm} joint velocity history       &   \textbf{36} \\
                 & joint target history        & 24             & joint target history         &   24 \\
                 & CPG phase information       & 13             & CPG phase information        &   13 \\ \hline
Exteroception    & height samples              & 208            &     height samples           &   208    \\ \hline
Privileged info. & contact states              & 4              &      contact states          &   4       \\
                 & contact forces              & 12             &      contact forces          &   12      \\
                 & contact normals             & 12             &      contact normals         &   12      \\
                 & friction coefficients       & 4              &      friction coefficients   &   4       \\
                 & thigh and shank contact     & 8              &      thigh and shank contact &   8       \\
                 & external forces and torques & 6              &      external forces and torques & 6        \\
                 & airtime                     & 4              &      airtime                     & 4        \\
                 & &               &      \textbf{payload mass}                         &  \textbf{1} \\ \hline
\end{tabular}
\label{tab:proprioceptive}
\end{table*}

\renewcommand{\thefigure}{S\arabic{figure}}
\setcounter{figure}{0}

\clearpage

\begin{figure*}
    \centering
    \includegraphics[width=\textwidth]{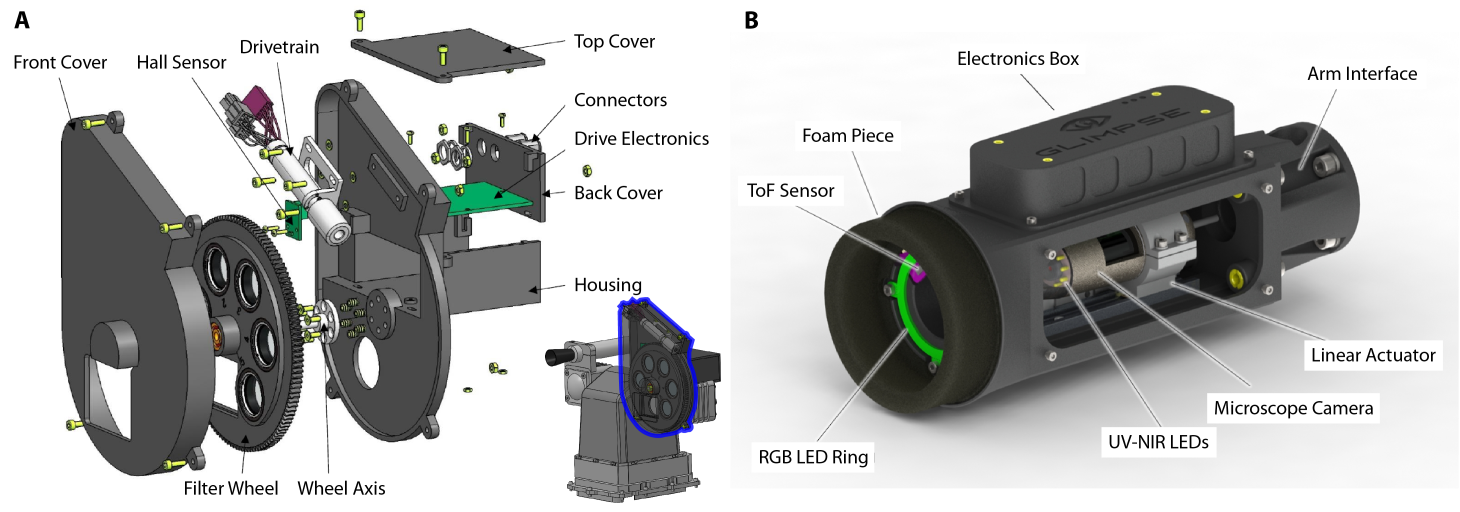}
    \caption{\textbf{Custom-built scientific payloads as described in Sec.~Science Payloads.} (\textbf{A}) Exploded view of the filter wheel, which is used on the Scout's CTX. (\textbf{B}) CAD rendering of MICRO with an open instrument panel.}
    \label{fig:ScientificPayloads}
\end{figure*}

\begin{figure*}
    \centering
    \includegraphics[width=\textwidth]{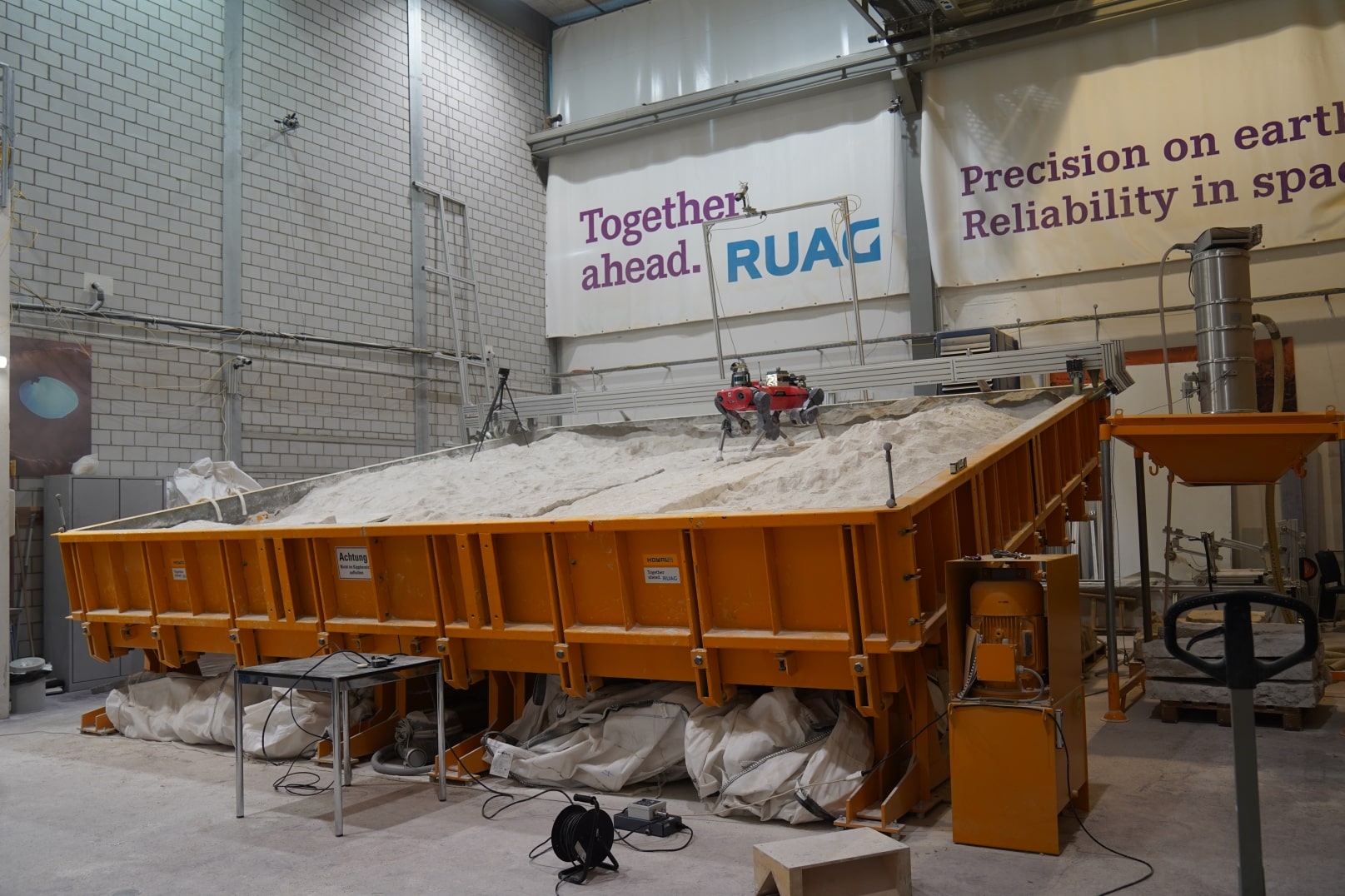}
    \caption{\textbf{ExoMars locomotion testbed at Beyond Gravity (former RUAG Space).} The testbed can be filled with planetary analog soil and tilted up to \SI{25}{\degree}. The respective test results are shown in Sec.~Locomotion Results.}
    \label{fig:LocomotionTestbed}
\end{figure*}

\begin{figure*}
    \centering
    \includegraphics[width=\textwidth]{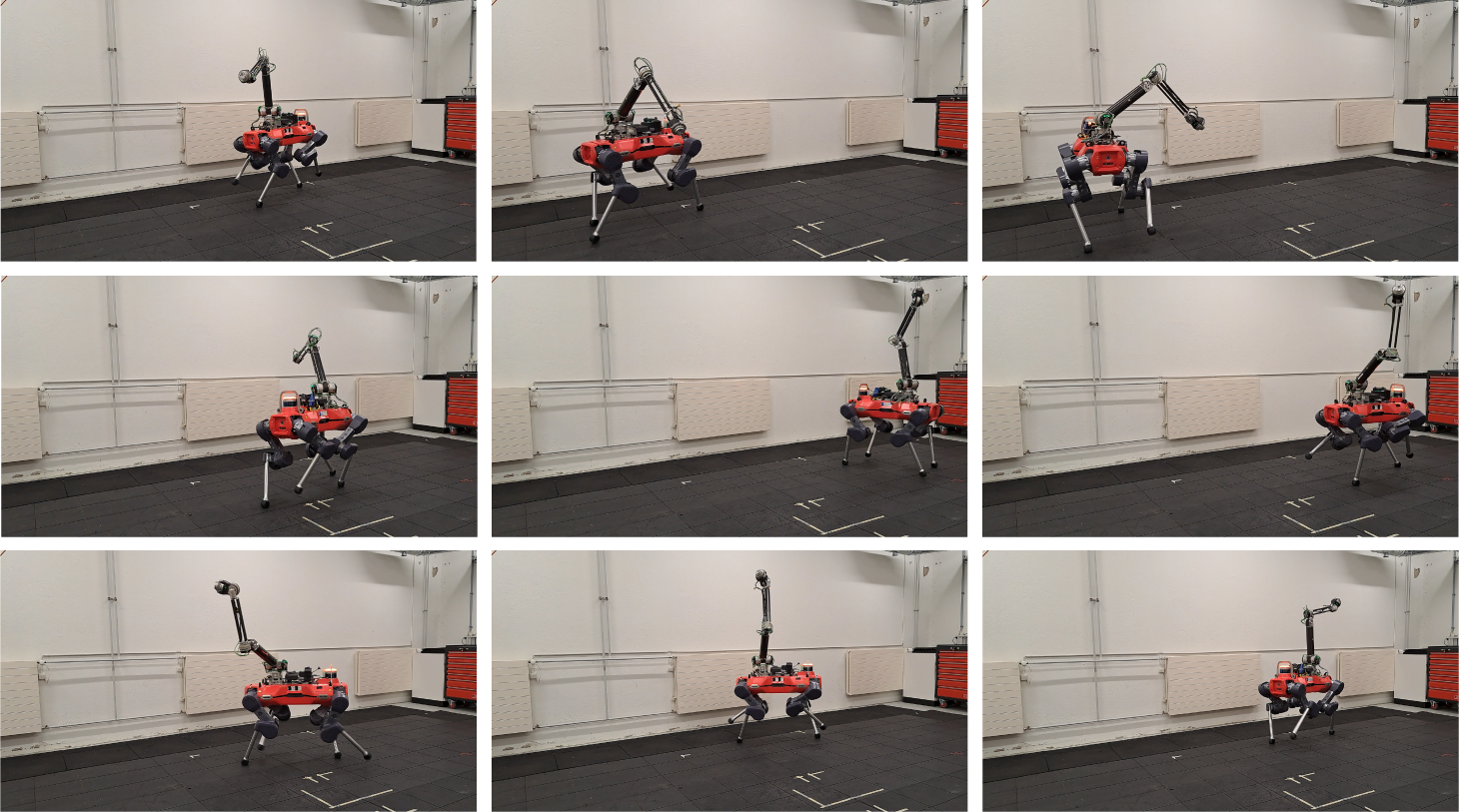}
    \caption{\textbf{Locomotion of the Scientist robot with a moving arm.} The image sequence demonstrates continuous and robust locomotion of ANYmal walking while the arm moves to randomly sampled set points.}
    \label{fig:LocomotionWithArm}
\end{figure*}

\begin{figure*}
    \centering
    \includegraphics[width=\textwidth]{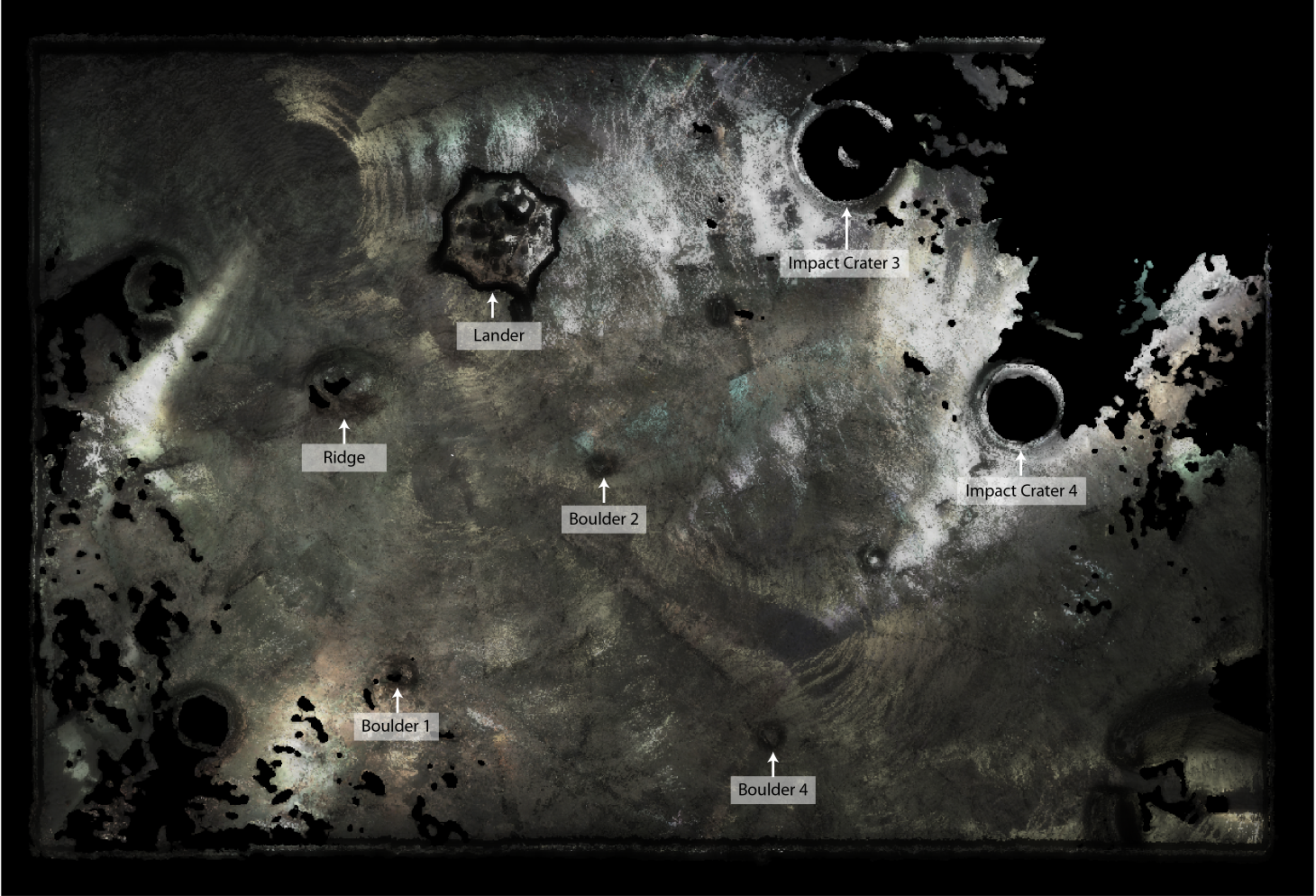}
    \caption{\textbf{Colorized dense mesh from the SRC analog mission.} Important landmarks are highlighted with labels similar to those shown in Fig.~3. }
    \label{fig:colorizedMapOfSpaceResourceChallenge}
\end{figure*}

\begin{figure*}
    \centering
    \includegraphics[width=\textwidth]{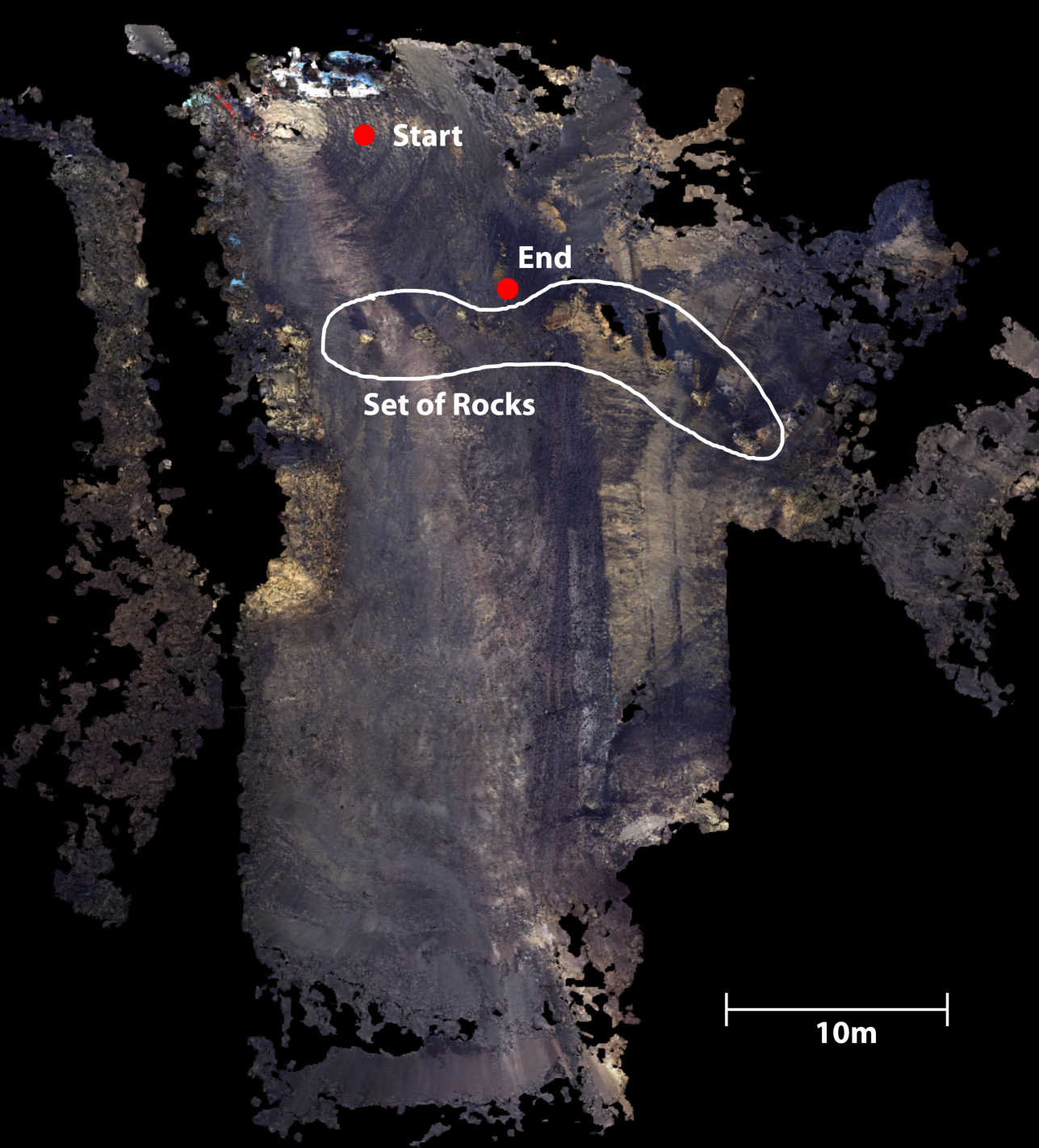}
    \caption{\textbf{Colorized dense mesh from the quarry analog mission.} The rock landmarks are highlighted, as well as the start and end positions. Fig.~S6 shows a generated mesh instance around the set of rocks.}
    \label{fig:colorizedMapOfNeuheim}
\end{figure*}

\begin{figure*}
    \centering
    \includegraphics[width=\textwidth]{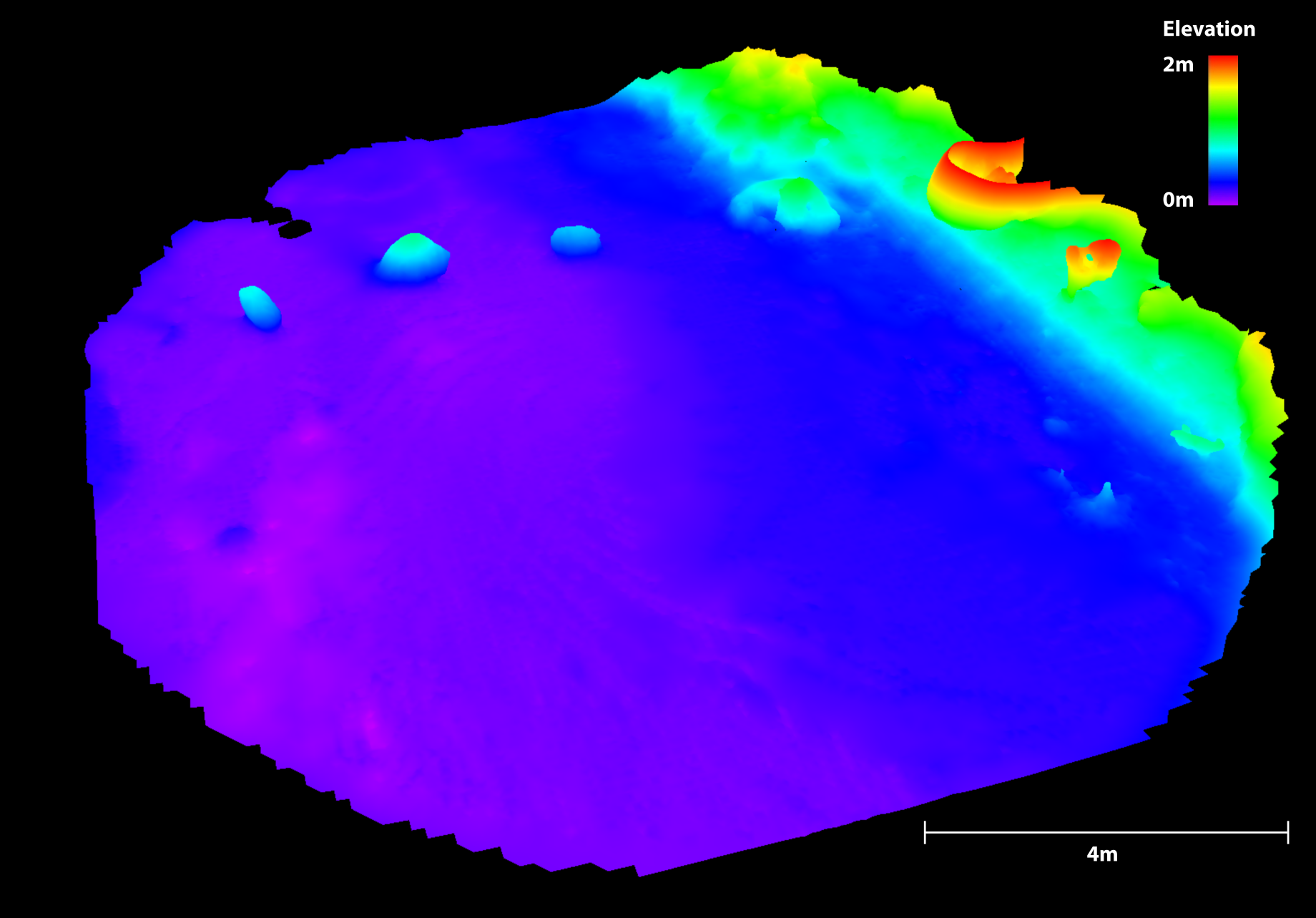}
    \caption{\textbf{Single online mesh instance from the quarry analog mission.} Colorization indicates the relative elevation of the vertices. Red indicates high elevation, whereas purple indicates low elevation. The rock landmarks shown in Fig.~S5 are clearly visible in the mesh. We describe the mesh generation in Sec.~Lightweight Mesh Representation}
    \label{fig:singleOnlineMeshInstance}
\end{figure*}

\begin{figure*}
    \centering
    \includegraphics[width=\textwidth]{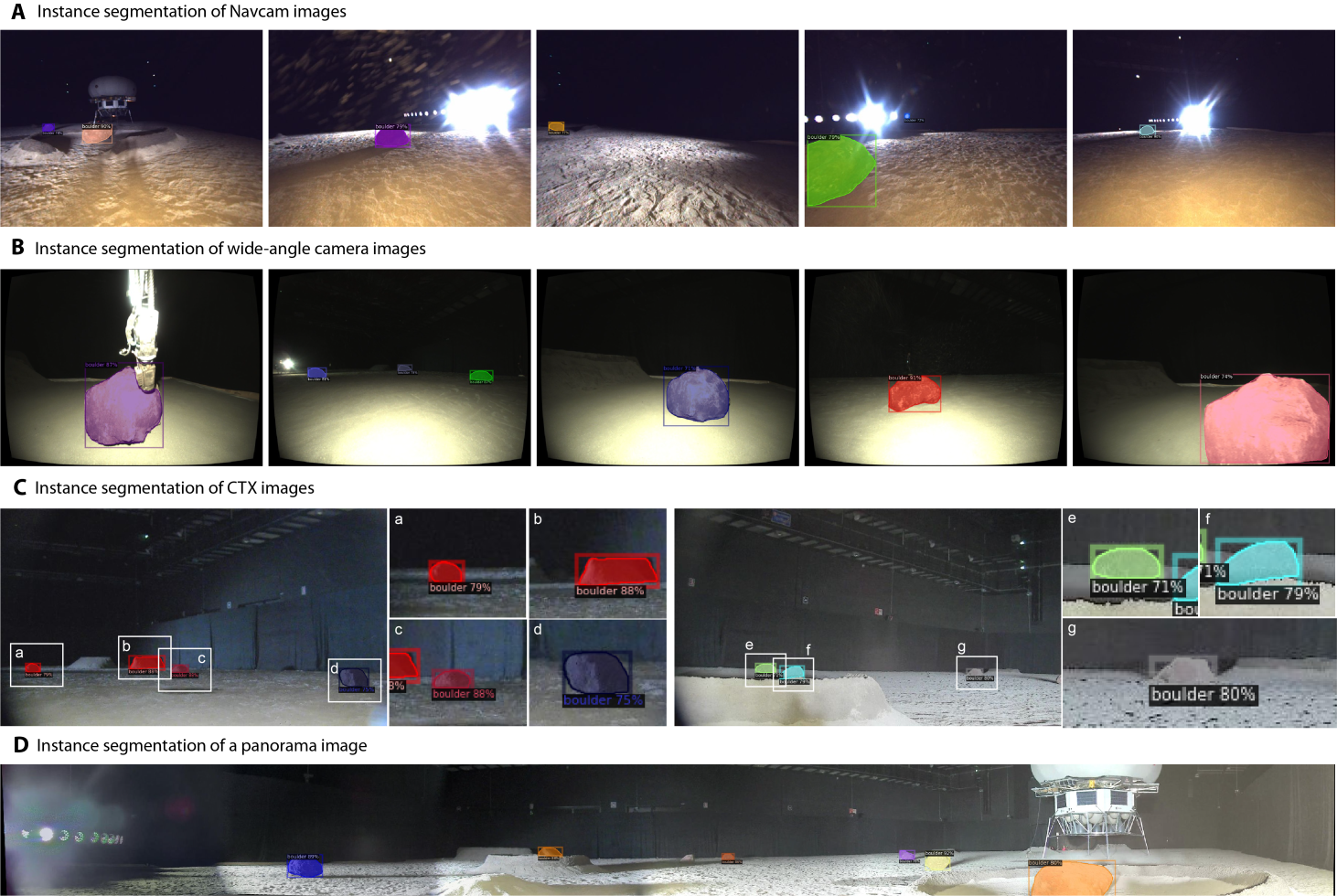}
    \caption{\textbf{Boulder instance segmentation results of images from different cameras.} (\textbf{A}) Instance segmentation results of Alphasense Navcam images. Masks and bounding boxes of different rock instances were labeled in different colors. (\textbf{B}) Instance segmentation results of wide-angle camera images. (\textbf{C}) Instance segmentation results of CTX images. Subfigures \textbf{a-g} are enlarged views of the identified rock instances labeled with masks, bounding boxes, and the corresponding confidence. (\textbf{D}) Instance segmentation result of a panorama image. Our segmentation pipeline is detailed in Sec.~Instance Segmentation}
    \label{fig:InstanceSegmentationResults}
\end{figure*}

\begin{figure*}
    \centering
    \includegraphics[width=\textwidth]{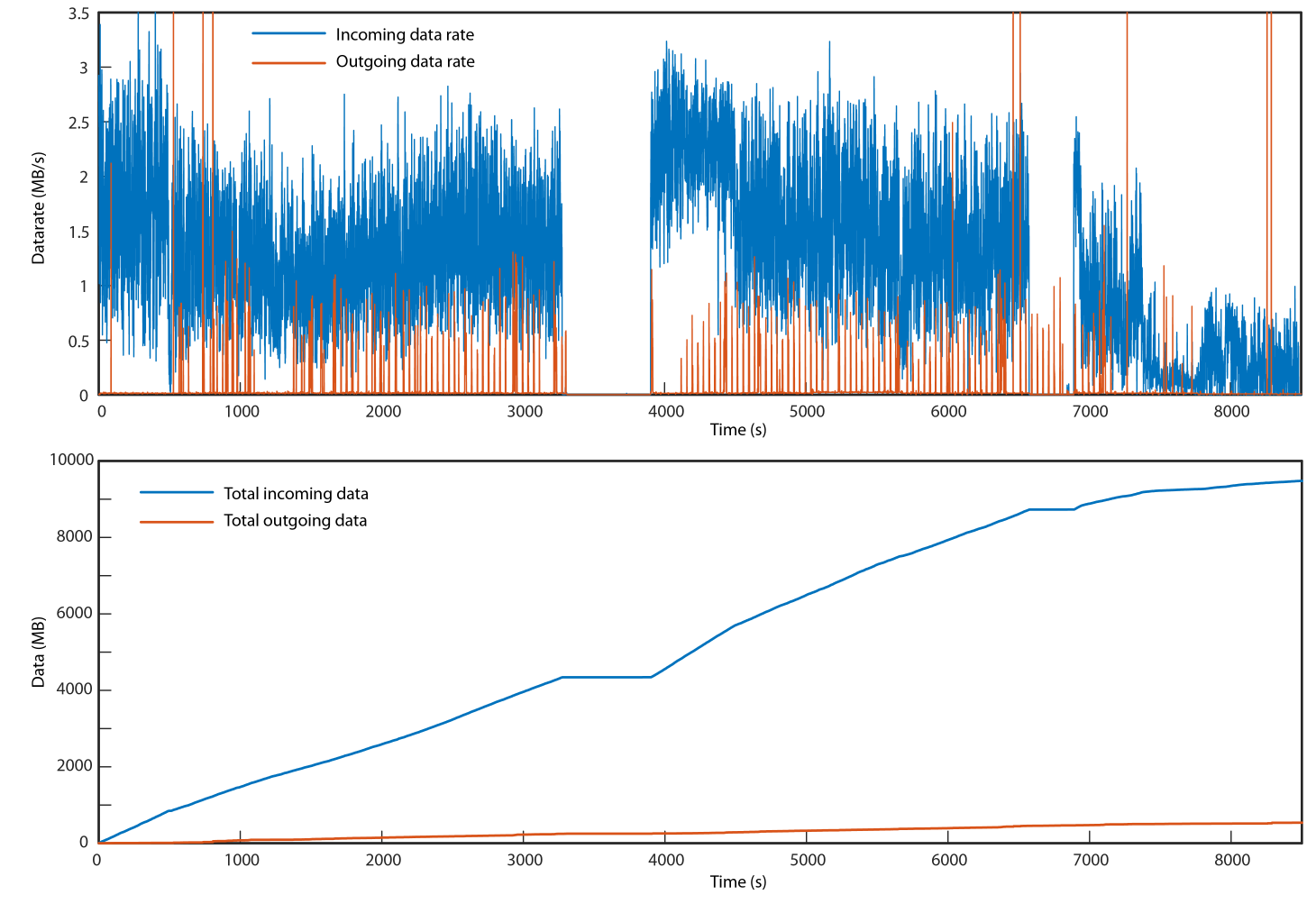}
    \caption{\textbf{Network data traffic on mission control PC 1 during the SRC.} The top figure shows the data rate during the mission and the bottom figure shows the accumulated transferred data. During the two LoS events, mission control received no data from the robot. Generally, the incoming data rate stays below \SI{3.5}{\mega \byte \per \second}.}
    \label{fig:mission_control_data_usage}
\end{figure*}

\begin{figure*}
    \centering
    \includegraphics[width=0.5\textwidth]{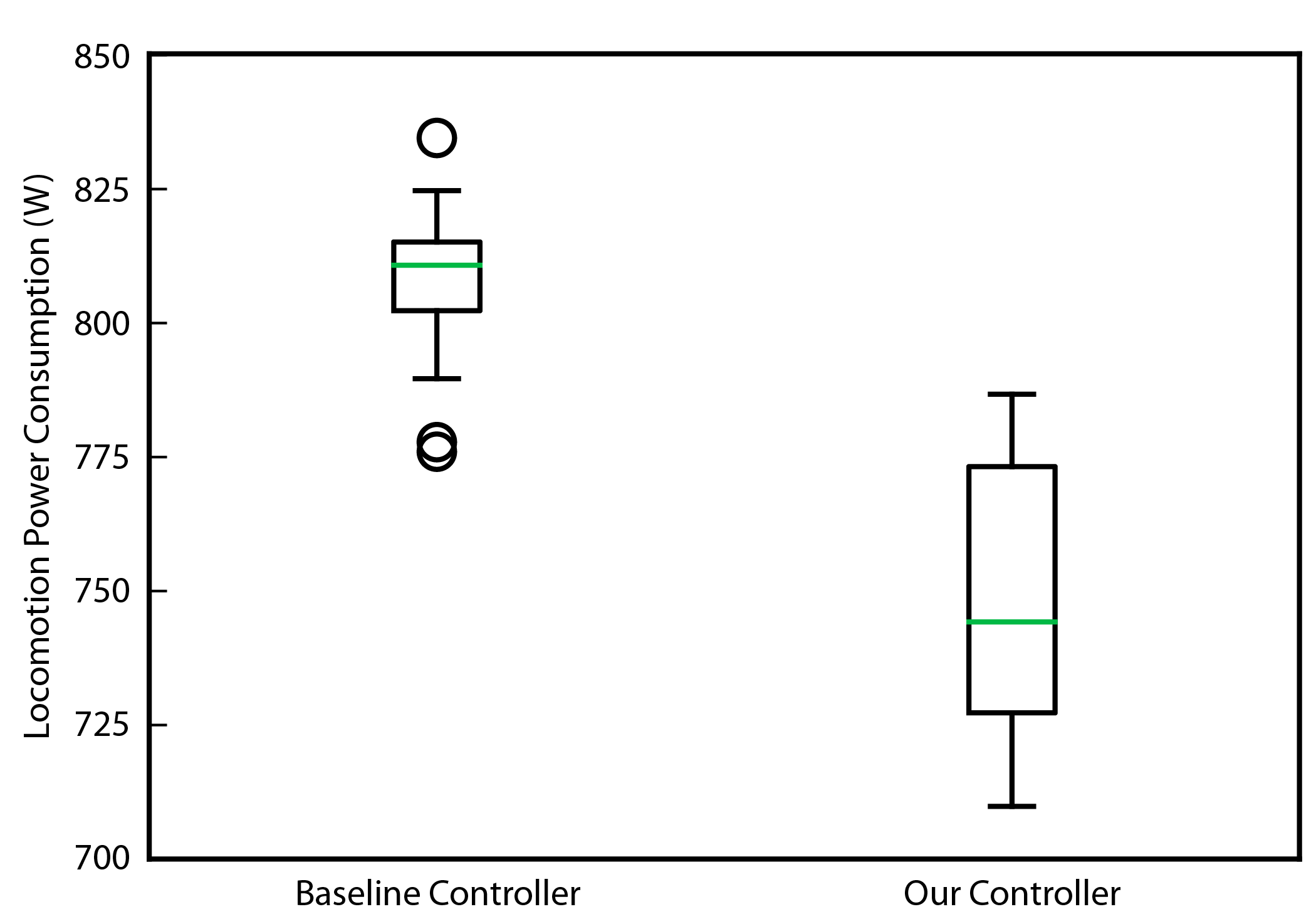}
    \caption{\textbf{Power consumption of our locomotion controller compared to the baseline \cite{miki2022learning}.} The visualized power consumption is the difference between the total power consumption of the robot during walking with the respective controller and the standby power consumption of the robot (\SI{175}{\watt}). Mann-Whitney U test; N = 40 for each controller, $P < 0.001$.}
    \label{fig:PowerConsumption}
\end{figure*}

\begin{figure*}
    \centering
    \includegraphics[width=\textwidth]{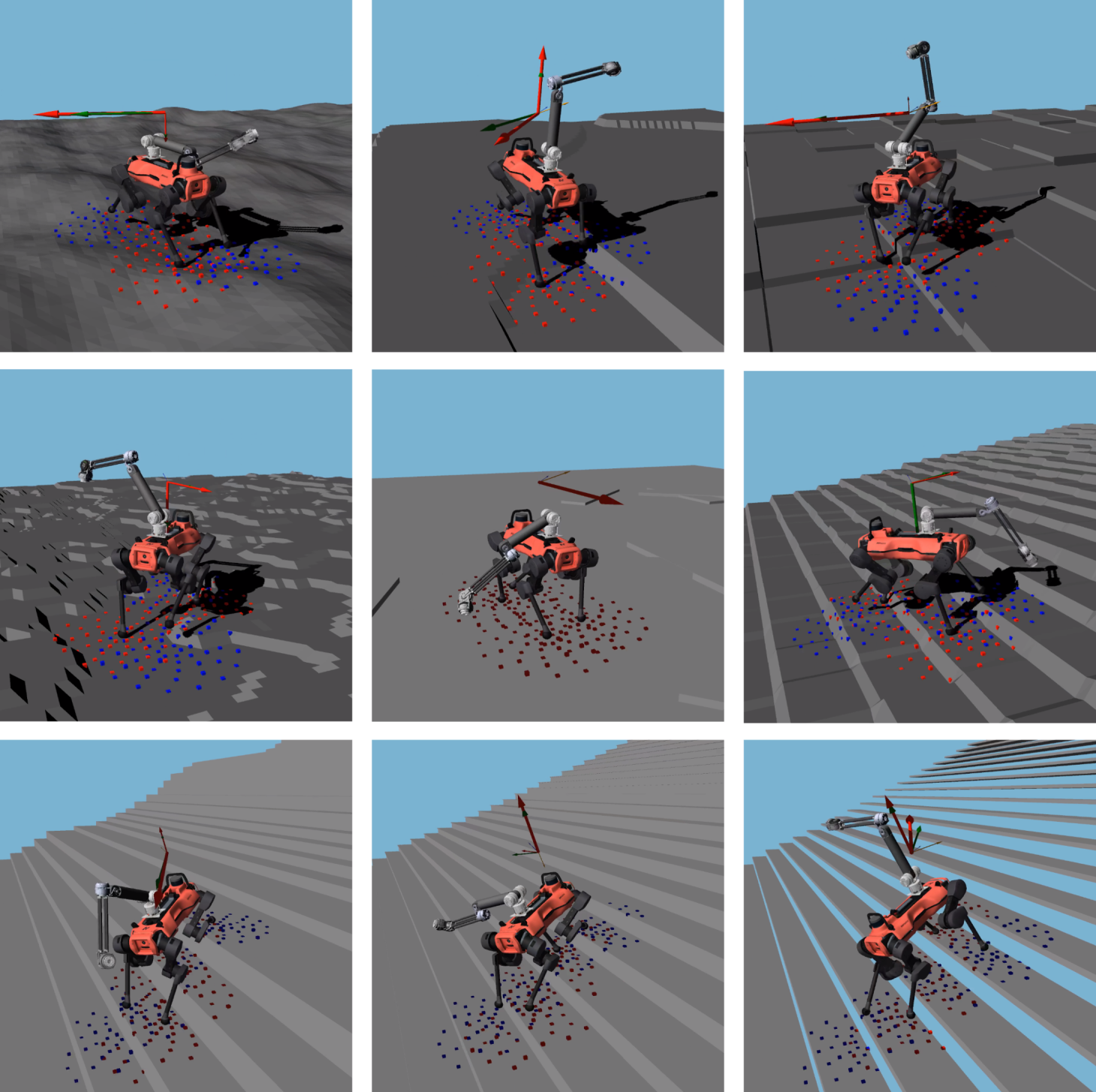}
    \caption{\textbf{Training environment of locomotion with arm.} The terrain parameters (for example slope angle and step height) are modified based on an adaptive curriculum to gradually increase the difficulty during training. The red arrows above the robot indicate the commanded velocity, and the green arrows represent the current velocity. The red and blue dots below the feet are the height scan providing information of the terrain's geometry. As seen in the images, the arm position is randomized during training.}
    \label{fig:LocomotionTraining}
\end{figure*}

\begin{figure*}
    \centering
    \includegraphics[width=0.8\textwidth]{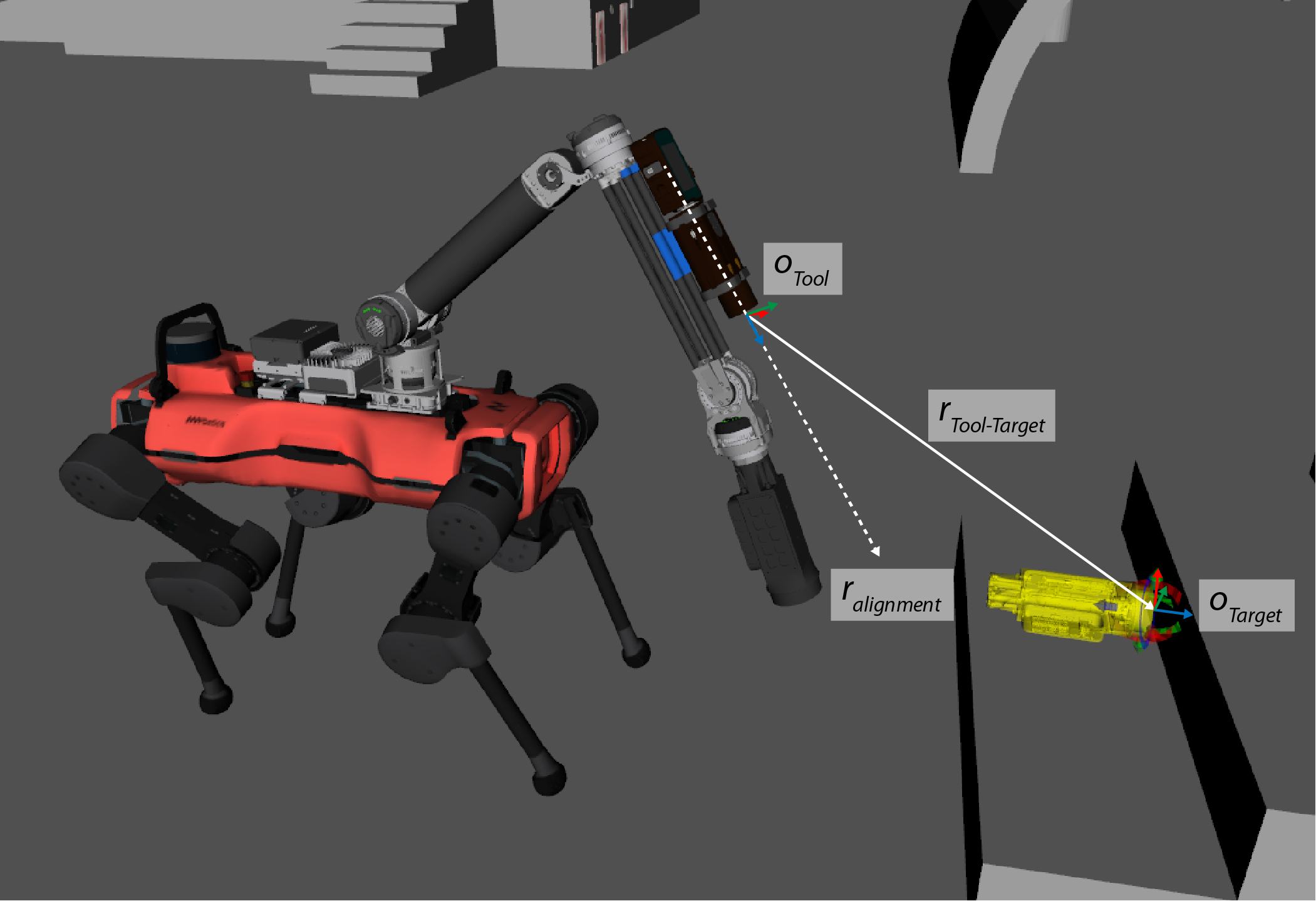}
    \caption{\textbf{Arm control frames for scientific payload deployment.} The target frame corresponds to the 6D pose set by the operator and the tool frame corresponds to the scientific payload.}
    \label{fig:MiraDeployment}
\end{figure*}

\begin{figure*}
    \centering
    \includegraphics[width=\textwidth]{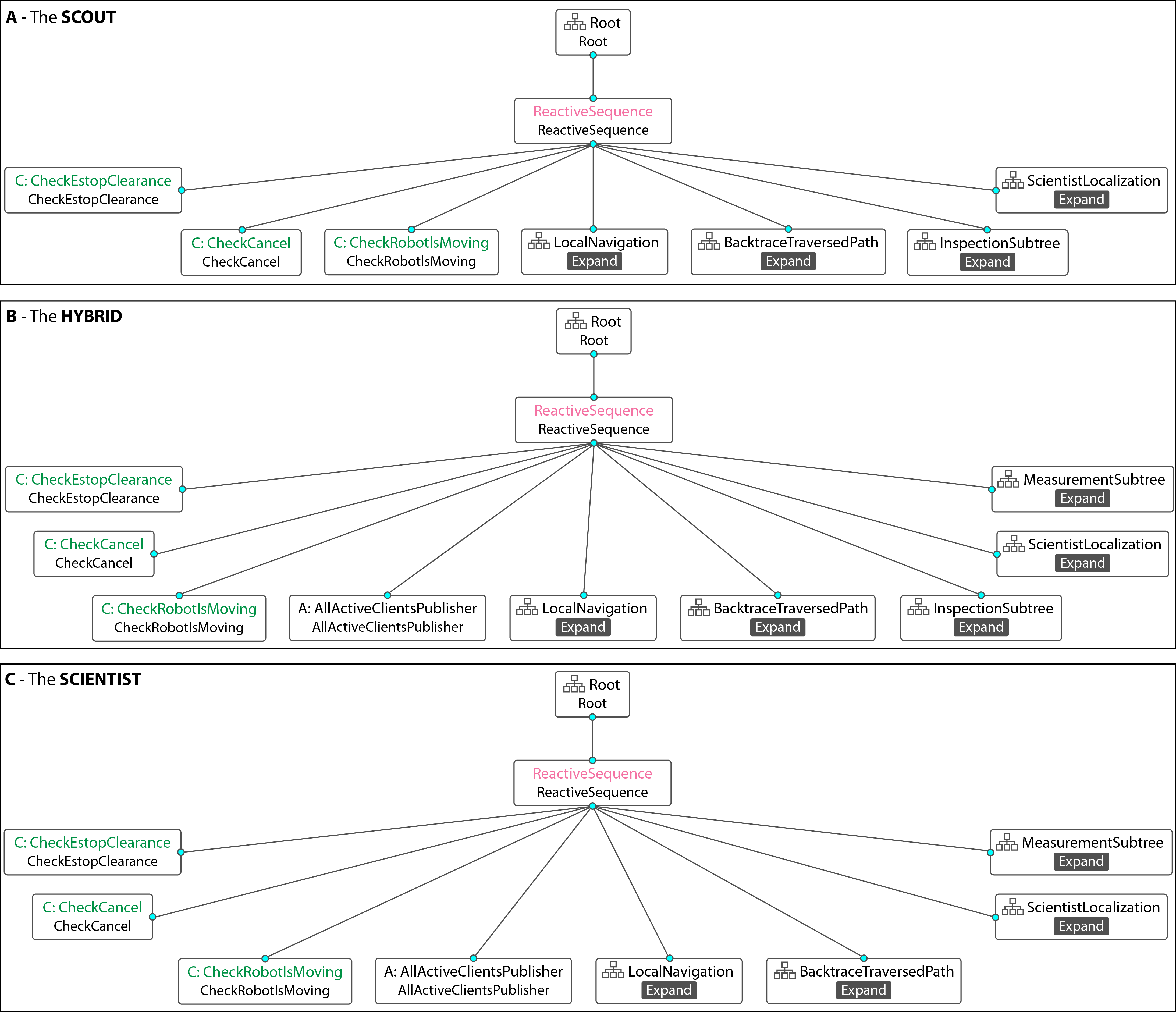}
    \caption{\textbf{An overview of the behavior trees used on each robot.} (\textbf{A}) The behavior tree of the Scout. (\textbf{B}) The behavior tree of the Hybrid. (\textbf{C}) The behaviour tree of the Scientist.}
    \label{fig:AllRobotBT}
\end{figure*}

\begin{figure*}
    \centering
        \label{fig:AllSubtree}
        \includegraphics[width=\textwidth]{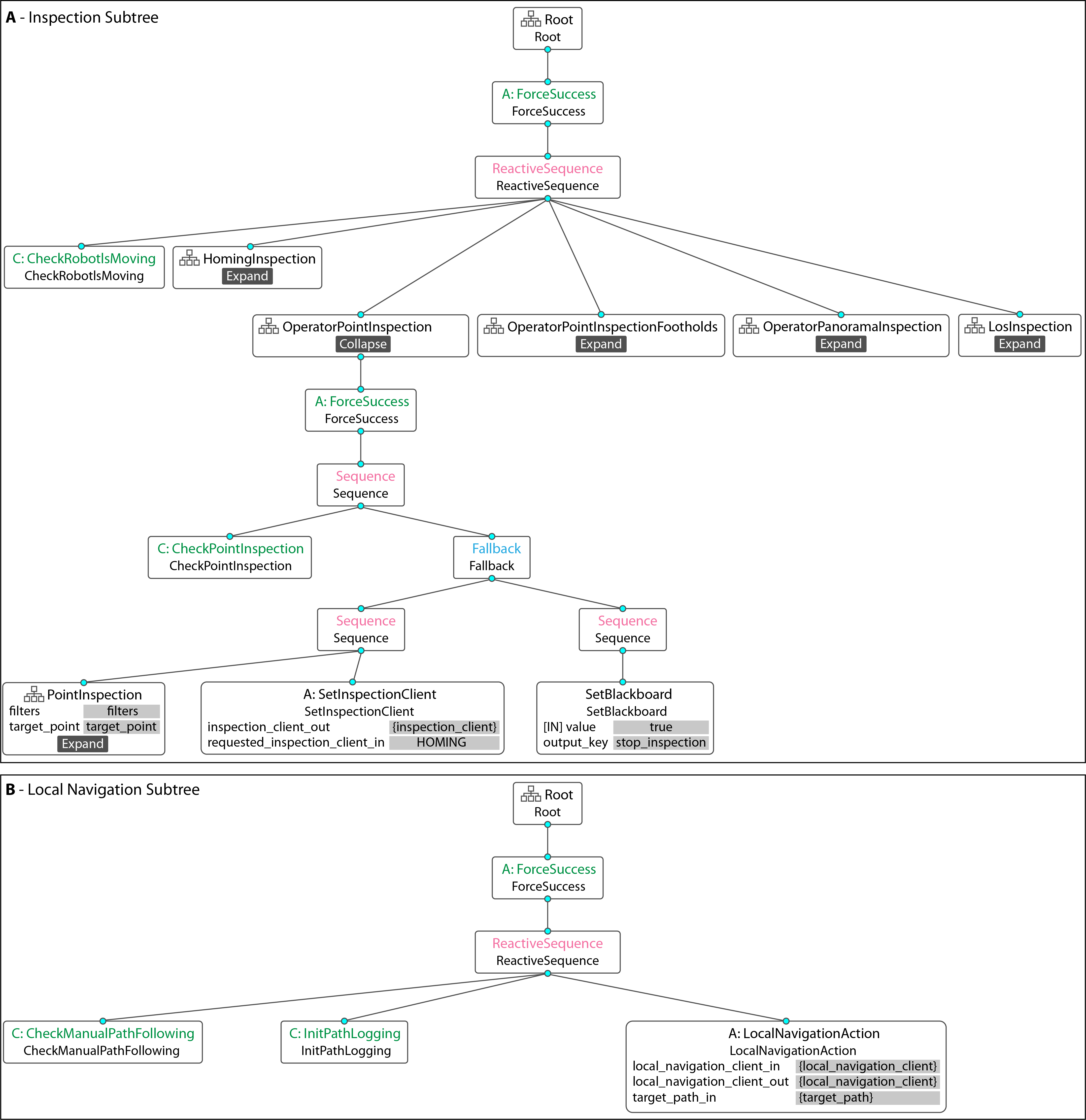}%
    \label{fig:AllSubtree}
\end{figure*}

\addtocounter{figure}{-1}
\begin{figure*}
    \addtocounter{figure}{1}
    \centering
        \includegraphics[width=\textwidth]{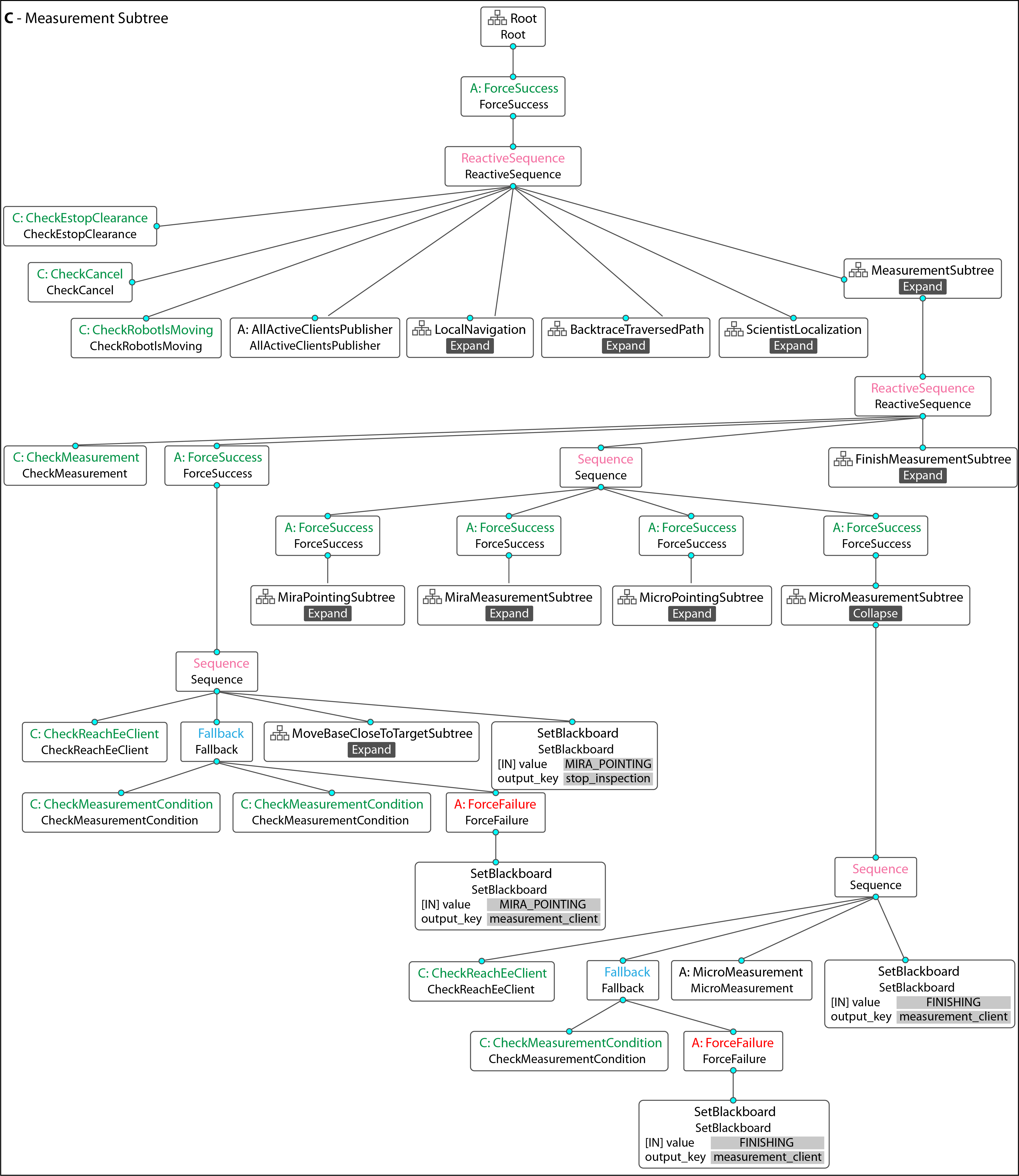}%
    \caption{\textbf{An overview of the most important subtrees used in the behavior trees for the three robots.} (\textbf{A}) The inspection subtree used in the behavior trees for the Scout and the Hybrid. (\textbf{B}) The local navigation subtree used in the behavior trees for all three robots. (\textbf{C}) The measurement subtree used in the behavior trees for the Hybrid and the Scientist.}
\end{figure*}

\end{document}